\documentclass[lettersize,journal]{IEEEtran}
\usepackage{amsmath,amsfonts}
\usepackage{algorithmic}
\usepackage{algorithm}
\usepackage{array}
\usepackage[caption=false,font=normalsize,labelfont=sf,textfont=sf]{subfig}
\usepackage{textcomp}
\usepackage{stfloats}
\usepackage{url}
\usepackage{verbatim}
\usepackage{graphicx}
\usepackage{array}
\usepackage{multirow}
\usepackage{cite}
\usepackage[table]{xcolor}
\begin{document}

\title{FFINet: Future Feedback Interaction Network\\ for Motion Forecasting}

\author{\IEEEauthorblockN{Miao Kang,
Shengqi Wang, Sanping Zhou,~\IEEEmembership{Member, IEEE}, Ke Ye, Jingjing Jiang, and
Nanning Zheng,~\IEEEmembership{Fellow, IEEE}}
\thanks{This work was supported by National Natural Science Foundation of China (No. 62088102, No. 62106192). ({\it Corresponding author: Nanning Zheng.})}
\thanks{Miao Kang, Shengqi Wang, Sanping Zhou, Ke Ye, Jingjing Jiang, and
Nanning Zheng are with the National Key Laboratory of Human-Machine Hybrid Augmented Intelligence, National Engineering Research Center for Visual Information and Applications, and Institute of Artificial Intelligence and Robotics, Xi'an Jiaotong University, Shaanxi 710049, China (E-mails: \{kangmiao, w867111858, yk123456\}@stu.xjtu.edu.cn, spzhou@xjtu.edu.cn, jingjingjiang2017@gmail.com,nnzheng@mail.xjtu.edu.cn).}}

\markboth{IEEE TRANSACTIONS ON INTELLIGENT TRANSPORTATION SYSTEMS}%
{Kang \MakeLowercase{\textit{et al.}}: FFINet: Future Feedback Interaction Network for Motion Forecasting}

\maketitle

\begin{abstract}
Motion forecasting plays a crucial role in autonomous driving, with the aim of predicting the future reasonable motions of traffic agents. Most existing methods mainly model the historical interactions between agents and the environment, and predict multi-modal trajectories in a feedforward process, ignoring potential trajectory changes caused by future interactions between agents. In this paper, we propose a novel Future Feedback Interaction Network (FFINet) to aggregate features the current observations and potential future interactions for trajectory prediction. Firstly, we employ different spatial-temporal encoders to embed the decomposed position vectors and the current position of each scene, providing rich features for the subsequent cross-temporal aggregation. Secondly, the relative interaction and cross-temporal aggregation strategies are sequentially adopted to integrate features in the current fusion module, observation interaction module, future feedback module and global fusion module, in which the future feedback module can enable the understanding of pre-action by feeding the influence of preview information to feedforward prediction. Thirdly, the comprehensive interaction features are further fed into final predictor to generate the joint predicted trajectories of multiple agents. Extensive experimental results show that our FFINet achieves the state-of-the-art performance on Argoverse 1 and Argoverse 2 motion forecasting benchmarks. 
\end{abstract}


\section{Introduction}
\IEEEPARstart{G}{iven} the observation information of agents, such as pedestrians and vehicles, motion forecasting~\cite{helbing1995social,varadarajan2022multipath++, zeng2021lanercnn, shi2021sgcn} aims to predict their plausible future positions, which is vital for the safety of intelligent autonomous systems. 
As various complicated factors (\textit{e.g.}, complex interactions and negotiations between multiple agents) account for the future motion of agents in dynamic traffic scenarios, motion forecasting becomes an extremely challenging task in practice.

\begin{figure}[t]
	\centering

	\includegraphics[width=0.47\textwidth]{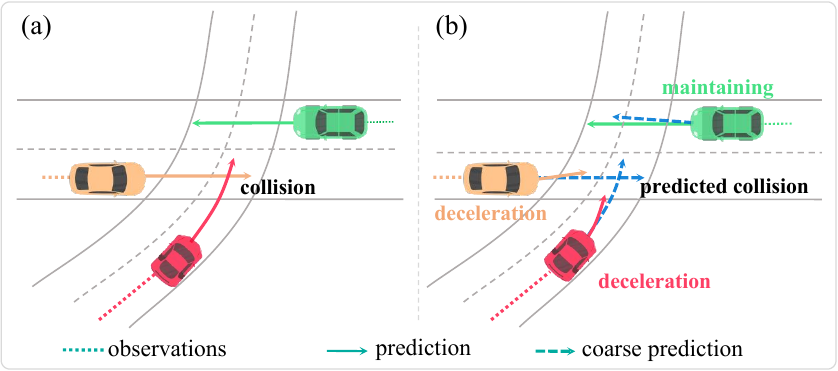}
	\caption{Motivation of FFINet. ~\textbf{(a)}~Traditional methods often generate incompatible trajectory predictions.
	~\textbf{(b)}~Human drivers generate feedback control to revise their feedforward control.}
	\label{fig:collision}
\end{figure}

Recent methods~\cite{gao2020vectornet, gu2021densetnt, li2020end} mainly focus on how to comprehensively and properly embed complex interactions with different models based on observation trajectories and environmental information. Even though significant progress has been achieved in motion forecasting, we argue that those methods are still hard to process the trajectory refinement based on the possible future interactions between different agents. As shown in Fig.~\ref{fig:collision}~(a), they forecast the motion of a single agent in a single feedforward pass and construct its correlation relationships with surroundings independently, which leads to the model ignoring the possible future collisions between predictions of agents. 

To alleviate those possible conflicts in the future, some prior works~\cite{tang2019multiple,casas2020implicit} formulated motion forecasting as a simple joint prediction problem, which allows penalizing the incompatibility of future trajectories. Some other methods~\cite{sun2022m2i, song2022learning} consider introducing future information into the feedforward prediction. For example, goal-based methods~\cite{gilles2022gohome, gu2021densetnt} often predict the agents' destinations and condition the distribution of future trajectories on the predicted goals.Although the future uncertainty is reduced with predicted goals, it still has multiple different paths to a certain goals, which is hard for the models to capture the possible interaction among the predictions of agents.
\IEEEpubidadjcol
In practice, human drivers solve this problem in a different way. As shown in Fig.~\ref{fig:collision}~(b), according to the historical information of neighbors and the current topology structure of a scene, they usually predict the agents' motion using an internal dynamics model to assess possible risks. Those initial assessments in turn help them generate feedback controls, such as acceleration and deceleration, to correct the discrepancies introduced by their feedforward control~\cite{nash2016review}. The key to human success lies on the assessments and feedback of possible future interaction between future trajectories, which are further utilized to revise the current predictions.
\label{sec:iEquippedntro}

Motivated by human drivers, we aim to explore the feedback effect of future interactions to enable the understanding of pre-action for motion forecasting. Compared with the general pipeline of motion forecasting methods, we decompose this feedback process into three critical steps:~(1) predicting the initial trajectories of all agents of each scene in a single feedforward pass according to the historical information and current topology structure of the scene;~(2) exploring the possible future interactions across agents according to the predicted trajectories; and~(3) feeding back the resulting future interactions into the feedforward pass to refine the initial prediction.

In this paper, we design a simple yet effective Future Feedback Interaction Network~(FFINet) for motion forecasting. In particular, the future interactions obtained from the initial joint predictions are fed back to the observation interaction, in which three kinds of interactions,~\emph{i.e.}, feedback interaction, future interaction and feedforward interaction, are introduced to refine the initial prediction results. The whole pipeline of our FFINet can be simply divided into the following three steps. Firstly, we decompose the representation of agents and HD maps into position vectors, current positions, and relative relationships. The position vectors and current positions are encoded to extract rich spatial-temporal features for cross-temporal aggregation. Secondly, with a relative interaction block handling the relative relationships, the cross-temporal aggregation propagates the information between current, observation and future interaction features. Thirdly, feeding into global fused features, we effectively refine the initial predictions with the final predictor. We conduct extensive experiments on both Argoverse 1 and Argoverse 2 motion forecasting benchmarks, in which our FFINet has achieved the state-of-the-art performance.

The main contributions of this paper can be summarized as follows:
\begin{itemize}
    \item We propose a novel motion prediction framework, in which the historical, current and future interaction features are integrated to understand pre-actions.
    \item We propose a novel cross-temporal aggregation strategy, in which a future feedback module is designed to explore the feedback effect of future interaction.
    \item We disentangle the inherent and relative motion of scene with data decomposition, which makes relative relationships can be explored in a relative interaction block.
\end{itemize}
The rest of this paper is organized as follows:
We briefly review the related work in Section~\ref{Related Works}. 
We present the technical details of our proposed method in Section~\ref{Our Method}. 
Then, extensive experiments and analysis are presented in Section~\ref{Experimental Analysis}. 
Finally, we conclude the paper in Section~\ref{Conclusion}.

\section{Related Work}
\label{Related Works}
\textbf{Interaction Models for Motion Forecasting.}
Complex interactions between agents and their surroundings have been the focus of motion forecasting.
Early researches~\cite{helbing1995social, antonini2006discrete} mainly represent interactions by hand-crafted energy functions, which cannot capture the rich behavioral strategies in the interactions. Recent approaches~\cite{varadarajan2022multipath++, gilles2022gohome} wildly utilize deep neural networks, including Recurrent Neural Network (RNN)~\cite{messaoud2019relational}, Graph Convolutional Network (GCN)~\cite{shi2021sgcn} and Conditional VAE(CVAE)~\cite{lee2022muse}, to depict the social and environmental interactions for agents. Social LSTM~\cite{alahi2016social} introduces a social pooling layer to learn typical human-human interactions automatically. Evolvegraph~\cite{li2020evolvegraph} captures the future dynamic interactions between agents with an evolution interaction graph.
VectorNet~\cite{gao2020vectornet} presents a hierarchical graph neural network to model the high-order interactions among all road components with vector representation.
LaneGCN~\cite{liang2020learning} exploits a fusion network to model four types of interactions between agents and lane segments, which is wildly employed by following researches~\cite{zhou2022hivt, zeng2021lanercnn}.
Since those interaction methods only model the observation interactions over individual agents and fail to reason about their interactions in the future~\cite{sun2022m2i}, they always produce incompatible future trajectories over multiple agents.  
Our FFINet introduces a future feedback module to capture the future interaction with initial predictions and performs cross-temporal interactions for comprehensive information extraction.

\textbf{Multi-agent Joint Motion Forecasting.}
To predict trajectories of multiple agent in a future scene simultaneously, some prior works~\cite{casas2020implicit, zhou2022hivt, girgis2021latent} learn a joint predictor to predict trajectories in a joint space over multiple agents. MPF~\cite{tang2019multiple} introduces a probabilistic framework that efficiently learns latent variables to model the multi-step future motions of agents jointly. 
Sergio et al.~\cite{casas2020implicit} propose an implicit latent variable model to characterize the joint distribution over socially consistent future trajectories. 
SceneTransformer~\cite{ngiam2021scene} formulates an attention-based encoder-decoder architecture for the joint prediction of all agents, producing consistent futures that account for interactions between agents. 

Although those models perform joint prediction, they typically select one central agent and normalize all other agents’ information based on this central agent, which harms the performance of models~\cite{jia2022multi}. Therefore, Kofinas et al.~\cite{kofinas2021roto} propose local coordinate frames pernode-object to introduce roto-translation invariance to the geometric graph of the interacting dynamical system. Jia et al.~\cite{jia2022multi} propose a symmetric
way of encoding the spatial relationship between elements in
the scene, which enables the model to predict multiple agents’ future in one forward without performance drop. 
HiVT~\cite{zhou2022hivt} addresses the joint prediction problem by decomposing the absolute coordinates into position vectors and relative information. 
To predict the initial trajectories of multi-agents in a scene, our FFINet also decomposes the data into position
vectors, current positions and relative relationship for comprehensive future interaction extraction.

\begin{figure*}[t]

\centering
\includegraphics[width=0.98\textwidth]{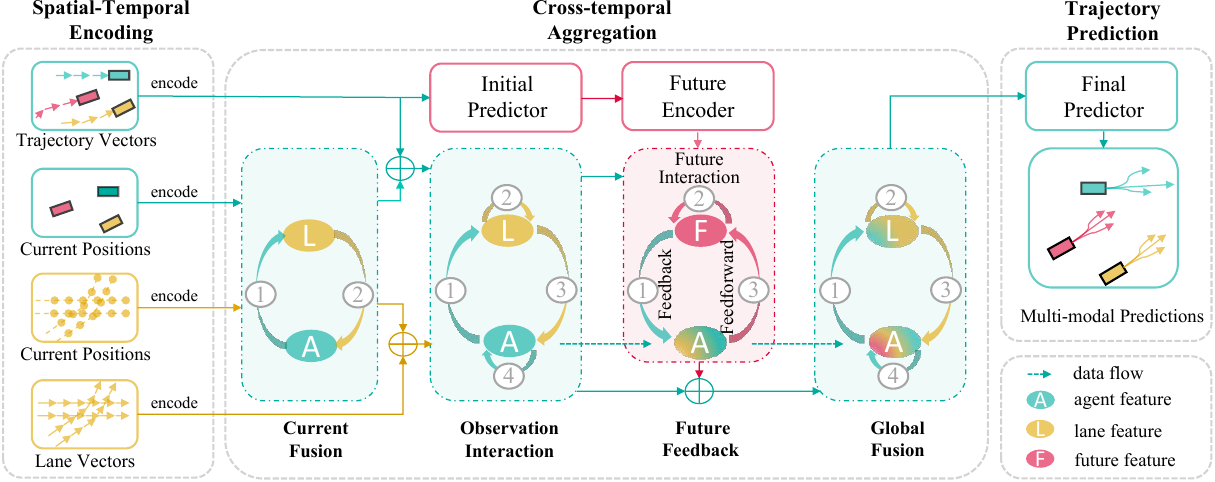} 
\caption{{Framework of FFINet}, in which:~\textbf{(1)} position vectors and current positions are embedded by separated encoders; ~\textbf{(2)} four interaction modules are introduced to aggregate cross-temporal interaction features;~\textbf{(3)} numbers with circle represent the order to perform interactions.}
\label{fig:overview}
\end{figure*}

\textbf{Conditioned Motion Forecasting.}
For introducing the future information into models, MTR~\cite{shi2022motion} first predicts the future trajectory of each agent, and concatenates the features of historical and future trajectories as representation of motion embedding. While they ignore the interactions between future trajectories, which makes it hard to deal with the possible future situation between multiple agents. The conditioned prediction methods~\cite{tolstaya2021identifying,mangalam2020not} generally forecast the future distribution of agents conditioned on the planning paths~\cite{song2022learning} or predicted goals~\cite{wang2023ganet,gu2021densetnt, xu2022remember}.
After explicitly distinguishing the influencer and reactor, M2I~\cite{sun2022m2i} predicts the trajectories of the influencer and conditions them to the prediction of the reactor. In practice, predicting the relationships of interactive agents is always tricky in a complex traffic environment. ScePT~\cite{chen2022scept} generates the prediction on the next timestamp for each agent via GRUs, which is further fed into the state vector with latent variable, repeatedly. Obviously, the next timestamp prediction naturally limits the model to capture the future interaction possibilities between different agents. Y-net~\cite{mangalam2021goals} estimates an explicit probability distribution over the agent’s long term goals and a few chosen waypoints and paths for modeling the uncertainty in future trajectories. PRIME~\cite{song2022learning} utilizes self-attention to explore the interaction between redundant planning paths of each agent for path selection. However, the future interactions between agents stay unexplored in those methods. Therefore, our FFINet introduces a future feedback module to propagate the future interaction between agents back to observation interaction features and further revise the feedforward predictions.

\section{Future Feedback Interaction Network}
\label{Our Method}

\subsection{Overview}
Most recent works~\cite{liang2020learning, zhou2022hivt} devote to exploring the observation interactions and predict trajectories for a single agent in a single feedforward pass. In contrast, we design a FFINet for trajectory predictions by aggregating the current, historical and future interaction features.To obtain the inputs of FFINet, we decompose the absolute coordinates of trajectories and lanes. As shown in Fig.~\ref{fig:overview}, the architecture of our FFINet is consists of: 
~(1) \textbf{Spatial-Temporal Encoding} is employed to embed the position vectors and current positions of a scene, providing rich motion features for the following cross-temporal aggregation. ~(2) \textbf{Cross-temporal Aggregation} integrates the features subsequently with a current fusion module, observation interaction module, future feedback module and global fusion module. The Current Fusion Module fuses the current position features within the nearest agent-lane pairs. They are further added with corresponding vector features and integrated into the Observation Interaction Module. The Future Feedback Module is equipped with a future interaction, a feedback interaction and a feedforward interaction. It propagates the information between future features and observation interaction features of agents.
The Global Fusion Module integrates the cross-temporal interaction features of agents and lanes.
~(3) \textbf{Final Predictor} predicts joint multi-modal trajectories of multiple agents in a single feedforward pass with regression branches and a classification branch.

\subsection{Data Decomposition}
\begin{figure}[t]
	\centering
	\includegraphics[width=0.475\textwidth]{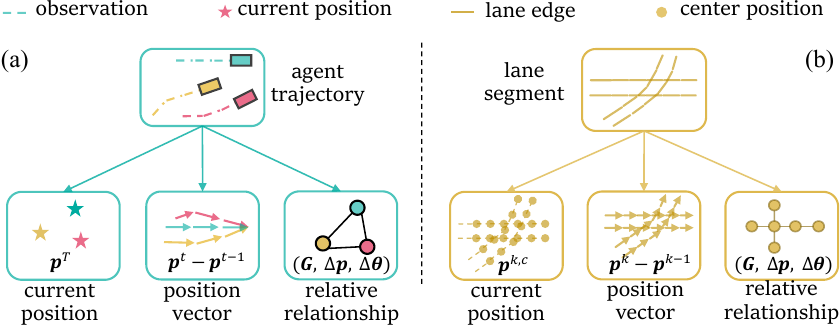}
	\caption{Multi-agent data decomposition of a scene.~\textbf{(a)}~Decomposition the trajectories of agents.~\textbf{(b)}~Decomposition the coordinates of lane segments.}
	\label{fig:dataprocess}
\end{figure}

To disentangle the inherent motion of agents and the relative correlation of a scene, as shown in Fig.~\ref{fig:dataprocess}, we decompose the absolute coordinates of trajectories and lanes into position vectors, current positions and relative relationship.

 \textbf{Position Vectors.} To represent inherent motion dynamics of agents, we vectorize the past trajectory of an agent $i$ with the position vectors $\mathbf{v}^t_i = \{\mathbf{p}^t_i - \mathbf{p}^{t-1}_i\}^T_{t=1}$, where $\mathbf{p}^t_i \subset \mathbb{R}^2 $ is the 2D coordinate of agent $i$ at timestamp $t$. Besides, $t \in [0, T]$ and $t \in [T+1, T+T_{p}]$ represent observation timestamps and predicted timestamps, respectively.
Likewise, for a lane segment $\xi$, we represent it as the position vector $\mathbf{v}^k_\xi = \{\mathbf{p}^{k}_\xi - \mathbf{p}^{k-1}_\xi\}_{k=1}^N$ and the center coordinate $\mathbf{p}^{k,c}_\xi = 0.5*(\mathbf{p}^{k}_\xi + \mathbf{p}^{k-1}_\xi)$, where $\mathbf{p}^{k}_\xi$ denotes the $k^{th}$ coordinates in $N$ points on lane segment $\xi$. 

 \textbf{Current Positions.} To enhance the current topology structure of a scene and extract the current fusion features, we preserve the current absolute coordinates $\mathbf{p}^{T}_i$ and $\mathbf{p}^{k,c}_\xi$ of agent $i$ and lane segment $\xi$ to preserve the initial relative position relationships of the scene. 

 \textbf{Relative Relationships.} To preserve relative spatial relationships between agents for interaction, we employ $\Delta\mathbf{p}_{ij} = \mathbf{p}^t_j - \mathbf{p}^t_i$ and $\mathbf{\theta}_{ij} = \mathbf{\theta}_j - \mathbf{\theta}_i$ to denote the relative distance and heading angle between agent $i$ and agent $j$ at timestamp $t$, where $\theta_i$ denotes the orientation of reference vector $\mathbf{p}^T_i - \mathbf{p}^{T-1}_i$. Similarly, $\Delta\mathbf{p}_{i\xi} = \mathbf{p}^T_i - \mathbf{p}^{k,c}_\xi$ and $\mathbf{\theta}_{i\xi} = \mathbf{\theta}_i - \mathbf{\theta}_\xi$ denote the relative distance and heading angle between agent $i$ and lane segment $\xi$. The relative relationships are employed by the relative interaction modules for correlation exploring.

In this way, the absolute coordinates of trajectories and lane segments can be fully restored from the decomposed components without any information loss, which allows our model to perform prediction over the entire future scene in a single forward pass efficiently.

\subsection{Spatial-Temporal Encoding}
With the decomposed data representation, we leverage separate encoders, including trajectory vector encoder, lane vector encoder and current position encoders, to embed position vectors and current positions of agents and lane segments.

\textbf{Trajectory Vector Encoder.} 
Given a scenario with multiple agents, we take the current absolute coordinates $\mathbf{p}^T_i$ of a given target agent $i$ as the origin of the coordinate system. Meanwhile, its heading angle $\theta_i$ is adopted as $x$-axis positive direction. We uniformly translate all position vectors of agents in this scenario to the origin and rotate them to the $x$-axis positive direction, which generates the normalized vectors as follows:
\begin{equation}
    \begin{aligned}
        \mathbf{z}^t_i &= { \mathbf{R}^\top_i(\mathbf{v}^t_i-\mathbf{p}^T_i)}, \\
    \mathbf{z}^t_j &={ \mathbf{R}^\top_j(\mathbf{R}^\top_i(\mathbf{v}^t_j-\mathbf{p}^T_i))},
    \label{eq:rotj}
    \end{aligned}
\end{equation}
where $\mathbf{R}^\top_i, \mathbf{R}^\top_j \subset \mathbb{R}^{2\times 2} $ are the  transpose of rotation matrix parameterized by $\theta_i$ and $\theta_j$.

With the normalized vectors, we firstly leverage a trajectory vector encoder with multi-scale 1D CNN layers to effectively extract the spatial features. Then, we further utilize a linear layer to gather the temporal information and obtain vector feature $\mathbf{e}^v_i$ as follows:
\begin{align}
    \mathbf{e}^v_i &={\operatorname{Linear}(\operatorname{Conv}([\mathbf{z}^t_i, \mathbf{a}_i]^T_{t=1}))},
    \label{eq:trajencode}
\end{align}
where $\mathbf{a}_i$ is a $1 \times T$ binary mask to indicate if the position vector at each step is padded or not. $[\cdot, \cdot]$ indicates the concatenation operation. 

\textbf{Lane Vector Encoder.} Similarly, in a lane vector encoder, we firstly normalize the position vector $\mathbf{v}^k_\xi$ of a lane segment $\xi$ and obtain the normalized vector $\mathbf{z}_\xi^k$ like Eq.(\ref{eq:rotj}).

Subsequently, we adopt the Lane Graph Convolution Network(LaneGCN) in~\cite{liang2020learning} to extract the topology information of lane segments, which can be formulated as follows:
\begin{align}
    \mathbf{e}^v_\xi &={\operatorname{LaneGCN}([\mathbf{z}^k_\xi, \mathbf{a}_\xi])},
    \label{eq:laneencode}
\end{align}
where $\mathbf{a}_\xi$ denotes the attributions of lane segments, including turning and traffic light information. 

\textbf{Current Position Encoder.} Current position encoders are utilized to embed the current 2D coordinates of agents and lanes with our MLP Block, which consists of two stacked blocks including a linear layer, a Norm layer and a ReLU function. We denote this MLP Block as $\operatorname{MLP}^1$ in the following sections. The current position encoder can be formulated as follows:
\begin{equation}
    \begin{aligned}
    \mathbf{z}^p_i &={\operatorname{MLP}^1(\mathbf{p}^T_i)},\\
    \mathbf{z}^{p}_\xi &={\operatorname{MLP}^1(\mathbf{p}^{k,c}_\xi)}.
    \label{eq:mapencode}
    \end{aligned}
\end{equation}

\subsection{Cross-Temporal Aggregation} 
With the vector and current features of the scene, we design a cross-temporal aggregation, including a current fusion module, an observation interaction module, a future feedback module and a global fusion module, to integrate the historical, current and future interaction features of multiple agents. In this section, we first introduce a relative interaction block, which is frequently reused by the observation interaction modules, future interaction modules and global fusion modules. Then we describe each module sequentially.

\begin{figure}[t]
	\centering
	\includegraphics[width=0.475\textwidth]{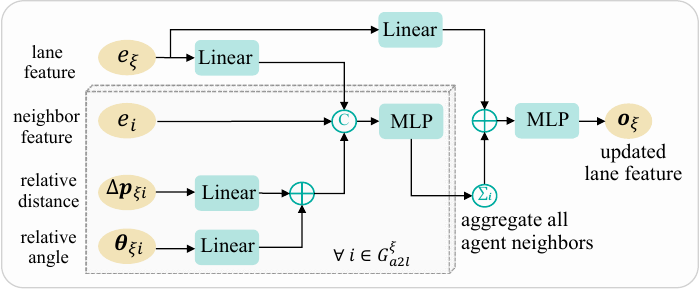}
	\caption{Illustration of Relative-correlation Interaction Block.}
	\label{fig:interaction}
\end{figure}

To explore the spatial correlations among agents and lanes with the decomposed relative information, we design a plug-in base block called \textit{relative interaction block}. As shown in Fig.~\ref{fig:interaction}, we take the interaction between lane segments and their surrounding agents as an example. Firstly, a neighbor agent's relative distance and angle are encoded and concatenated with the feature of this agent and lane segment. This process repeats on each neighbor agent of the lane segments. Secondly, we aggregate the concatenated features of all neighbors and update the raw feature of this lane segment with them.

Given a lane segment $\xi$ and its neighbor agents, this interaction process can be denoted as:
\begin{equation}
    \begin{aligned}
\mathbf{o}_\xi=\mathcal{I}(\mathbf{e}_\xi, \mathbf{e}_i, \Delta p_{\xi i}, \theta_{\xi i}), \forall i\in G^{\xi}_{a2l},
    \label{eq:block}
    \end{aligned}
\end{equation}
where $\mathbf{e}_\xi$ and $\mathbf{e}_i$ are the inherent motion feature of lane segment $\xi$ and a neighbor agent $i$. $G^{\xi}_{a2l}$ denotes a set of neighbor agents which connected with lane segment $\xi$ on graph $G_{a2l}$. The detailed process can be formulated as follows:
\begin{eqnarray}
    \mathbf{e}_{\xi i} \hspace{-0.2cm} &=&  \hspace{-0.2cm}{\operatorname{ReLU}(\operatorname{MLP}^2(\Delta p_{\xi i}) + \operatorname{MLP}^2(cos(\theta_{\xi i}), sin(\theta_{\xi i})))}, \nonumber\\
    \mathbf{o}_{\xi i} \hspace{-0.2cm} &=&  \hspace{-0.2cm}{\operatorname{MLP}^1( [\mathbf{e}_{\xi i}, \operatorname{MLP}^2({\mathbf{e}_\xi}),\mathbf{e}_i])}, \nonumber\\
    \mathbf{o}_\xi \hspace{-0.2cm} &=&  \hspace{-0.2cm} {\operatorname{MLP}^1(\operatorname{MLP}^2(\mathbf{e}_\xi) + \sum\nolimits_{i\in{G^{\xi}_{a2l}}}\mathbf{o}_{\xi i})},
\end{eqnarray}
where $\operatorname{MLP}^2$ represents the stack of two linear layers, a ReLU function.  

\textbf{Current Fusion Module.}
To make use of the current topology information of agents and lane segments, we further integrate the current feature between nearest agent-lane pairs, which can be denoted as follows:
\begin{equation}
    \begin{aligned}
    \mathbf{e}^{p_1}_{i} &={\operatorname{MLP}^1([\mathbf{z}^p_i, \mathbf{z}^{p}_{\xi i}])},\\
    \mathbf{e}^{p}_{i} &={\operatorname{MLP}^1([\mathbf{e}^{p_1}_{i}, \operatorname{MLP}^1(\mathbf{e}^{p_1}_{i}))]}.
    \label{eq:em3}
    \end{aligned}
\end{equation}

Similarly, we can obtain the current feature $\mathbf{e}^p_\xi$ for lane segment $\xi$. Finally, we add the current features to vector features and obtain the inherent motion features of agent $i$ and lane segment $\xi$, which can be formulated as follows:
\begin{equation}
    \begin{aligned}
        \mathbf{e}_i ={\mathbf{e}^v_i+\mathbf{e}^p_i},\\ \mathbf{e}_\xi ={\mathbf{e}^v_\xi+\mathbf{e}^p_\xi}.
    \label{eq:mapencode}
    \end{aligned}
\end{equation}

\textbf{Observation Interaction Module.}
The observation interaction module is leveraged to capture the observed relative correlation between agents and surroundings with the relative relationships and extracted features in Eq.(\ref{eq:mapencode}). Equipped with our relative interaction block, we follow LaneGCN~\cite{liang2020learning} to construct four types of interactions, including Agents to Lanes~(A2L), Lanes to Lanes~(L2L), Lanes to Agents~(L2A), and Agents to Agents~(A2A). According to Eq.(\ref{eq:block}), it can be formulated as:
\begin{equation}
    \begin{aligned}
    \text{A2L}:\mathbf{o}_\xi &=\mathcal{I}(\mathbf{e}_\xi, \mathbf{e}_i, \Delta p_{\xi i}, \theta_{\xi i}), \forall i\in G^{\xi}_{a2l},\\ 
    \text{L2L}:\mathbf{o}_\xi &=\phi(\mathbf{o}_\xi, \mathbf{o}_\zeta, \Delta p_{\xi \zeta}, \theta_{\xi \zeta}), \forall \zeta\in G^{\xi}_{l2l},\\
    \text{L2A}:\mathbf{o}_i &=\mathcal{I}(\mathbf{e}_i, \mathbf{o}_{\xi}, \Delta p_{i \xi}, \theta_{i \xi}), \forall \xi \in G^{i}_{l2a},\\
    \text{A2A}:\mathbf{o}_i &=\mathcal{I}(\mathbf{o}_i, \mathbf{o}_j, \Delta p_{i j}, \theta_{i j}), \forall j \in G^{i}_{a2a},\\
    \label{eq:obs}
    \end{aligned}
\end{equation}
where $\phi$ denotes the LaneGCN which has the same architecture of lane vector encoder.

\textbf{Future Feedback Module.}
To capture the possible future interactions and feed them back to the observations, we propose a future feedback module to exert the feedback influence of future interactions on the feed-forward prediction. Specifically, with the vector features of agents obtained from Eq.(\ref{eq:trajencode}), we first utilize an initial predictor to forecast a single trajectory prediction of each agent on future $T_p$ timestamps. Subsequently, a future encoder is used to embed those initial predictions, which can be formulized as,
\begin{equation}
    \begin{aligned}
        \mathbf{\hat{p}}^c_i &= \operatorname{Linear}(\operatorname{MLP}^1(\mathbf{e}^v_i)), \\
    \mathbf{e}^{_f}_i &= \operatorname{Linear}(\operatorname{Conv}(\mathbf{\hat{p}}^c_i)),
    \end{aligned}
\end{equation}
where $\mathbf{\hat{p}}^c_i$ indicates the predicted initial trajectory, and $\mathbf{e}^{_f}_i$ represents the future feature of agent $i$. The future encoder has the same structure as our trajectory vector encoder. After obtaining the future feature of agents, we integrate them with the corresponding observation interaction features obtained from Eq.(\ref{eq:obs}) with three types of interactions:

(\romannumeral1) \textit{Feedback Interaction.} We first integrate the observation interaction features of the target agent with the future features of corresponding surrounding agents. According to Eq.(\ref{eq:block}), this process can be denoted as $\mathbf{f}^b_i=\mathcal{I}^{(1)}(\mathbf{o}_i, \mathbf{e}_j^{_f}, \Delta p_{ij}, \theta_{ij})$, $\forall j\in G^i_{a2a}$,

where $G^i_{a2a}$ denotes all agents around the agent $i$ within a certain distance threshold, including the target agent itself. 

(\romannumeral2) \textit{Future Interaction.} We subsequently design a future interaction to explore the plausible interaction behaviors between agents, which are vital clues for feedback refinement. Similarly, it can be denoted as $\mathbf{f}^{u}_i=\mathcal{I}^{(2)}(\mathbf{f}^b_i, \mathbf{f}_j^b, \Delta p_{ij}, \theta_{ij})$, $\forall j\in G^i_{a2a}$.

(\romannumeral3) \textit{Feedforward Interaction.} We further update the observation interaction features to the future interaction features in a feedforward way, which forms a closed-loop information propagation between historical and future information. We obtain the feedforward interaction features $\mathbf{f}^{_f}_i=\mathcal{I}^{{(3)}}(\mathbf{f}^u_i, \mathbf{o}_j, \Delta p_{ij}, \theta_{ij})$, $\forall j\in G^i_{a2a}$. 

Finally, for a given agent $i$, we add the updated feedforward interaction feature $\mathbf{f}^{_f}_i$ up to the observation interaction features, so as to generate a cross-temporal fusion features for each agent as follows:
\begin{equation}
    \mathbf{f}_i = \mathbf{o}_i + \mathbf{f}^{_f}_i.
\label{eq:fi}
\end{equation}

\textbf{Global Fusion.} We introduce a simple global fusion module to propagate the context feature into the obtained cross-temporal features of each agent, which can provide a comprehensive representation of each agent for reasonable trajectory prediction. In particular, taking the observation lane feature $ \mathbf{o}_\xi$ in Eq.(\ref{eq:obs}) as well as the cross-temporal fusion feature $\mathbf{f}_i$ in Eq.(\ref{eq:fi}), we employ the same network architecture with the observation interaction module, and generate the final feature $\mathbf{g}_i$ for a given agent $i$.

\subsection{Final Prediction}
Taking the final feature $\mathbf{g}_i$ as input, our final predictor can generate normalized coordinates of multi-modal trajectories for multiple agents in a single shot. They can be transformed to absolute coordinates with the inverse operations in Eq.(\ref{eq:rotj}). Specifically, for providing $K$ possible future trajectories and their probabilities of best-predicted trajectory, we equip the prediction head with $K$ regression branches and a classification branch. The process can be formulated as follows:
\begin{equation}
    \begin{aligned}
            \mathbf{\hat{p}}^k_i &=\operatorname{MLP}^3(\mathbf{g}_i),\\ s^k_i&= \operatorname{MLP}^3([\operatorname{MLP}^4(\mathbf{\hat{p}}^{k,T_p}_i), \mathbf{g}_i)]),
    \end{aligned}
\end{equation}
where $\operatorname{MLP}^3$ consists of the MLP Block following a linear layer; and $\operatorname{MLP}^4$ is the stack of a linear layer, a ReLU function and a linear layer. Besides, $\mathbf{\hat{p}}^{k,T_p}_i$ denote the endpoint of the $k^{th}$ predicted trajectory of agent $i$.

\textbf{Model Training.} Unlike the two-stage goal-based methods~\cite{zhao2020tnt, gu2021densetnt}, our FFINet obtains the initial and final predictions in a single shot. Therefore, we calculate the error of initial and final predictions simultaneously. Specifically, we employ Smooth $l_1$ loss $\mathcal{L}^c_{reg}$ to measure the regression distance error between the initial predictions and ground-truth trajectories. As for the final multi-modal predictions, we take the winner-takes-all~(WTA)~\cite{lee2016stochastic} strategy and  back-propagate the loss of the best-predicted trajectory. Besides, we measure the regression loss  $\mathcal{L}_{reg}$ across all time steps, calculate the endpoint loss $\mathcal{L}_{end}$ with the smooth $l_1$ loss and the classification loss $\mathcal{L}_{cls}$ with the max-margin loss, as done in~\cite{liang2020learning}. The total loss  $\mathcal{L}$ can be denote as follows:
\begin{align}
    \mathcal{L} = \mathcal{L}_{reg} + \lambda \mathcal{L}_{end} + \beta \mathcal{L}_{cls} + \gamma \mathcal{L}^c_{reg},
\end{align}
where $\lambda$, $\beta$ and $\gamma$ are three constant weights. 

\begin{table*}[]
    \footnotesize 
    \centering
    \setlength{\tabcolsep}{4mm}
    \caption{Comparison with the state-of-the-art methods on Argoverse 1 Motion Forecasting Leaderboard.Numbers with bold fonts and underlines represent the best results and suboptimal results respectively.}

    \begin{tabular}{c|c|c|c|c|c|c|c|c}
    \hline
     \multirow{2}*{Methods} & \multirow{2}*{Year} &\hspace{-0.3cm} brier-minFDE~$\downarrow$ \hspace{-0.3cm} & minFDE~$\downarrow$ &\hspace{0.1cm} MR~$\downarrow$ \hspace{0.1cm} & minADE~$\downarrow$& minFDE~$\downarrow$ & \hspace{0.1cm} MR~$\downarrow$ \hspace{0.1cm} & minADE~$\downarrow$\\
     ~ & ~ & (K=6) & (K=6) & (K=6) & (K=6) & (K=1) & (K=1) & (K=1)\\
     \hline
     LaneGCN~\cite{liang2020learning} & 2020 & 2.054 & 1.362 & 0.162 &	0.870 & 3.762 & 0.588 & 1.702 \\
     DenseTNT~\cite{gu2021densetnt} & 2021 & 1.976 &	1.282 &	0.126 & 0.882 & 3.632 & 0.584 &	1.679\\
     TPCN~\cite{ye2021tpcn} & 2021 & 1.929 &	1.244 & 0.133 & 0.815 & 3.487 & 0.560 & \underline{1.575}\\
     Scene transformer~\cite{ngiam2021scene} & 2021 & 1.887 &1.232 & 0.126 & 0.803 & 4.055 & 0.592 & 1.811\\
     GOHOME~\cite{gilles2022gohome} & 2022 & 1.983 & 1.450 & {\color{black}\textbf{0.105}} & 0.943 & 3.647 & 0.572 & 1.689 \\
     HiVT-128~\cite{zhou2022hivt} & 2022 & 1.842 &	1.169 &	0.127 & 0.774 & 3.533 & \underline{0.547} & 1.598\\
     DSP\cite{zhang2022trajectory} & 2022 & 1.858 & 1.219 & 0.130 & 0.819 & 3.709 & 0.575 & 1.679 \\
     TENET~\cite{wang2022tenet} & 2022 & 1.839 & 1.229 & 0.124 &0.822 & 3.748 & 0.575 & 1.711\\
     Multipath++~\cite{varadarajan2022multipath++} & 2022 & 1.793 & 1.214 & 0.132 & 0.790	& 3.614 & 0.565 & 1.624\\
     GANet~\cite{wang2023ganet} & 2023 & 1.790 & \underline{1.161} & 0.118 & 0.806 & \underline{3.455} & 0.550 & 1.592\\
     Wayformer~\cite{nayakanti2023wayformer} & 2023 & \underline{1.741} & 1.162 & 0.119 & \underline{0.768} & 3.656 & 0.572 & 1.636\\
     \rowcolor{gray!20} FFINet & -- &	{\color{black}\textbf{1.729}} &	{\color{black}\textbf{1.121}} &	\underline{0.112}	&{\color{black}\textbf{0.761}}& {\color{black}\textbf{3.361}} &	{\color{black}\textbf{0.543}} & {\color{black}\textbf{1.533}}\\
        \hline   
    \end{tabular}
    \label{tab:Comparison}
\end{table*}

\begin{table}
    \footnotesize 
        \caption{Comparison on single-agent performance with the state-of-the-art methods on Argoverse 2 Leaderboard.}
    \centering
	\resizebox{0.475\textwidth}{!}{
    \begin{tabular}{c|c|c|c|c|c}
    \hline
         \multirow{2}*{Methods} & \multirow{2}*{Year} & \hspace{-0.25cm} brier-minFDE~$\downarrow$ \hspace{-0.25cm} & minFDE~$\downarrow$ & \hspace{0.2cm}MR~$\downarrow$\hspace{0.2cm} & minADE~$\downarrow$\\
         ~ & ~ & (K=6) & (K=6) & (K=6) & (K=6)\\
         \hline
         THOMAS~\cite{gilles2022thomas} & 2022 & 2.16 & 1.51 &0.20 & 0.88  \\
         GoRela~\cite{cui2023gorela} & 2023  & 2.01 &1.48 &0.22 &0.76\\
         FRM~\cite{park2023leveraging} & 2023 & 2.47 & 1.81 & 0.29 & 0.89\\
         GANet~\cite{wang2023ganet} & 2023 & \underline{1.96}  & \textbf{1.34}  & \textbf{0.17}& \underline{0.72} \\
         \rowcolor{gray!20} FFINet & -- & \textbf{1.95}& \underline{1.38}	& \underline{0.19}& \textbf{0.70} \\
        \hline
    \end{tabular}}
    \label{tab:av2}
\end{table}

\section{Experiments}
\label{Experimental Analysis}
\subsection{Experimental Settings}
\textbf{Dataset}.
Our FFINet is evaluated on the Argoverse 1 and Argoverse 2 Motion Forecasting Dataset, which contains high-definition~(HD) maps and trajectories of agents collected from real traffic scenarios. Argoverse 1 consists of 324,557 interesting agent trajectories, in which the first 2 seconds~($T=20$) of each trajectory are used as the observations, and the next 3 seconds~($T_p=30 $) are leveraged to supervise the predictions.
Argoverse 2 selects 250,000 scenarios for training and validation, where multiple high-quality trajectories relevant to AV are selected as "scored\_actors" in each scenario for multi-agent prediction. The first 5 seconds~($T=50$) of each trajectory are used as the observations, and the next 6 seconds~($T_p=60 $) are utilized for predictions.

\textbf{Evaluation Metrics.} We predict multi-modal trajectories~$(K=6)$ for each agent and evaluate our FFINet with single-agent and multi-agent metrics, respectively. As for single-agent evaluation, we adopt the standard metrics in Argoverse Motion Forecasting Competition~\cite{argoverse2022comp}, including minimum Final Displacement Error~(minFDE), minimum Average Displacement Error~(minADE), Miss Rate~(MR) and Brier minimum Final Displacement Error~(brier-minFDE). 
For multi-agent evaluation, we take the metrics of Argoverse2 Multi-World Forecasting  Challenge~\cite{argoverse22023comp}, including Average Brier Minimum Final Displacement Error (avgB-MinFDE), Average Minimum Final Displacement Error (avgMinFDE), Average Minimum Average Displacement Error (avgMinADE) and Actor Collision Rate (actorCR).

\begin{figure*}[t]
	\centering
	\includegraphics[width=1.0\textwidth, height=0.4\textwidth]{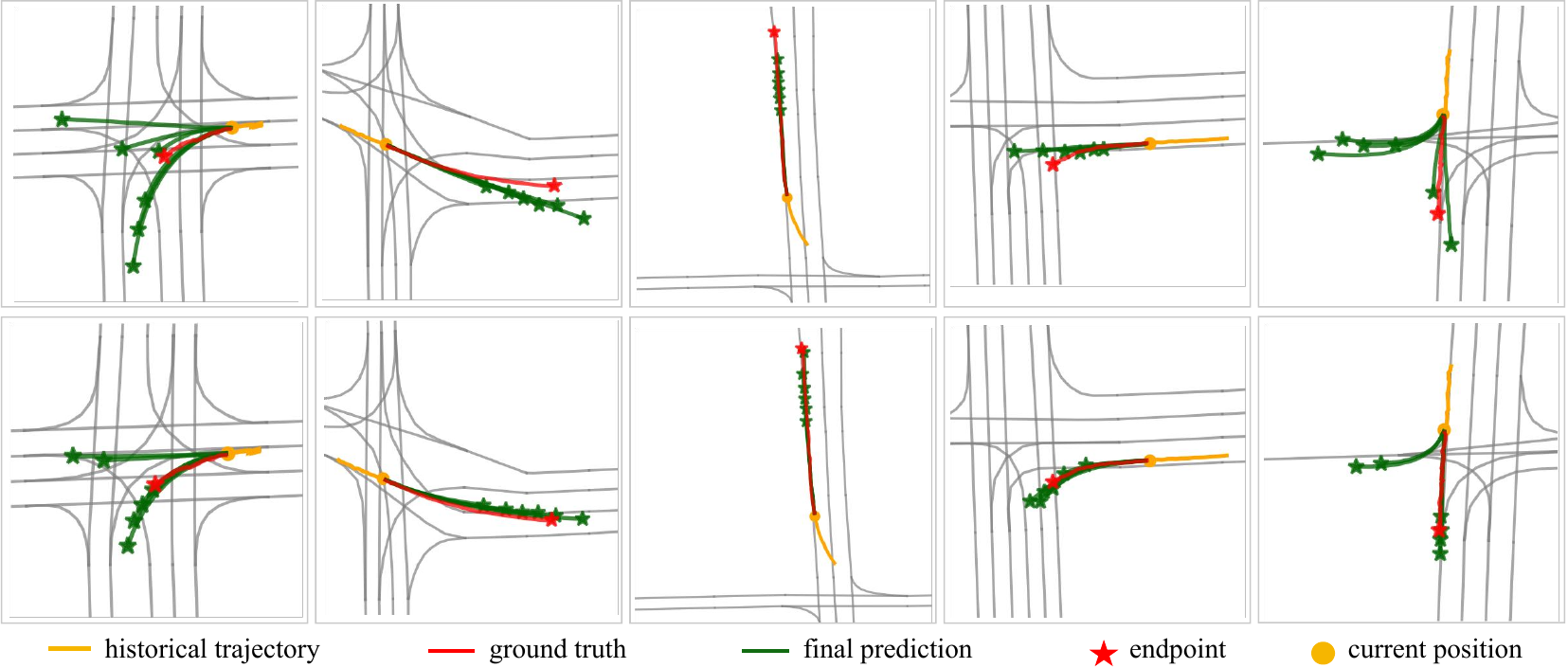}
	\caption{Results visualization of LaneGCN and FFINet on Argoverse validation set. The top row shows the predictions of LanGCN and the bottom raw presents the results of FFINet. The yellow, red, green and grey solid lines represent historical trajectory, ground truth trajectory, final predictions, and lanes respectively. For better visualization, we mark the endpoint of a trajectory with a star and the current position with a yellow point.}
	\label{fig:Comparison}
\end{figure*}

\textbf{Implementation Details.} The distance threshold of A2L, L2A and A2A are set to
7, 6, and 100 meters, respectively. The hidden size of the MLP Block is set as 128. In addition, the weight parameters in training loss are set as $\lambda=0.5, \beta=2.0, \gamma=0.5$, respectively. We take the Pytorch toolbox and two RTX 3090 Ti GPUs to train our model with a batch size of 64 for 40 epochs, which takes around 19 hours. Besides, we adopt the Adam optimizer with an initial learning rate of 1e-3, which is decayed to 1e-4 at 32 epochs following~\cite{liang2020learning}.
Since our model can simultaneously predict the trajectories of multiple agents, we just calculate the loss of trajectories longer than 40 timestamps for a balance between the quantity and quality of training data.
We didn't adopt any data augmentation method in this paper.

\begin{table}
    \footnotesize 
    \caption{Comparing on multi-agent performance with the state-of-the-art methods on the validation set of Argoverse 2.}
    \centering
    \resizebox{0.475\textwidth}{!}{
    \begin{tabular}{c|c|c|c|c}
    \hline
         Methods & Year & avgMinFDE~$\downarrow$ & actorMR~$\downarrow$ & avgMinADE~$\downarrow$\\
         \hline
         MTP~\cite{cui2019multimodal} &2019 &1.54 & 0.24 & 0.68 \\
         MultiPath~\cite{chai2020multipath} & 2020 & 2.13 & 0.33 &0.89 \\
         LaneGCN~\cite{liang2020learning} &2020  & 1.34 & 0.22 & 0.55 \\
         SceneTransformer~\cite{ngiam2021scene} &2021  &1.24 & 0.20 & \underline{0.52}\\
         GoRela~\cite{cui2023gorela} &2023 & \underline{0.96} & \underline{0.14} & \textbf{0.42} \\
         \rowcolor{gray!20} FFINet & -- & \textbf{0.93} & \textbf{0.11} & 0.55 \\
        \hline
    \end{tabular}}
    \label{tab:multi-agent}
\end{table}

\subsection{Comparison with State-of-the-Arts}
    \textbf{Comparison on Single-agent Performance.} We compare our method on the test set of Argoverse 1 with various state-of-the-art methods, including LaneGCN~\cite{liang2020learning}, DenseTNT~\cite{gu2021densetnt}, TPCN~\cite{ye2021tpcn}, Scene transformer~\cite{ngiam2021scene}, GOHOME~\cite{gilles2022gohome}, HiVT~\cite{zhou2022hivt}, DSP\cite{zhang2022trajectory}, TENET~\cite{wang2022tenet}, Multipath++~\cite{varadarajan2022multipath++}, GANet~\cite{wang2023ganet} and Wayformer~\cite{nayakanti2023wayformer}. For fair comparison, we further apply a simple yet effective ensemble strategy following~\cite{wang2022tenet} with the submissions on leaderboard. The results are shown in Table~\ref{tab:Comparison}, where our method consistently outperforms the other methods on all metrics except for MR~(K=6). Although we were narrowly defeated by GOHOME on MR~(K=6), we outperforms it in other six metrics by a large margin. It should be noticed that our method excels other methods by a large margin in brier-minFDE, ~\emph{i.e.}, over 18\% improvement than LaneGCN, which demonstrates its strong capability on both trajectory prediction and probability estimation \footnote{FFINet ranked $2^{th}$ on the Argoverse leaderboard on 25/02/2023.}.

To demonstrate the generalization of model, we also compare the single-agent performance of our FFINet with various state-of-the-art methods on the testing set of Argoverse 2 in Table~\ref{tab:av2} following the same strategy on Argoverse 1, including THOMAS~\cite{gilles2022thomas},  GoReLa~\cite{cui2023gorela}, FRM~\cite{park2023leveraging} and GANet~\cite{wang2023ganet}. FFINet achieves very competitive results compared with the newest works.

\textbf{Comparison on Multi-agent Performance.} As our model has the capability to predict multi-agents trajectories of each scenario simultaneously, we evaluate the performance of multi-agent prediction on the validation set of Argoverse 2 by following GoRela~\cite{cui2023gorela}. As shown in Table~\ref{tab:multi-agent}, comparing with the previous methods, including MTP~\cite{cui2019multimodal}, MultiPath~\cite{chai2020multipath},  LaneGCN~\cite{liang2020learning}, SceneTransformer~\cite{ngiam2021scene} and GoRela, FFINet significantly outperforms them on avgMinFDE and actorMR and achieves competitive results on avgMinADE with GoRela.

For more comprehensive evaluation on multi-agent prediction, we retrained the LaneGCN and HiVT on Argoverse2 and submited the tesing result of LaneGCN, HiVT and FFINet for the competition of multi-agent prediction on Argoverse2.
As shown in Table\ref{tab:av2leaderboard}, our FFINet outperforms the others on multi-agent metrics. Significantly, compared with LaneGCN, the Actor Collision Rate of FFINet decreased by 71.4\%, which demonstrates the effectiveness of future interaction feedback in collision avoidance.

\begin{table}
    \footnotesize 
    \centering
    \caption{Comparing on multi-agent performance with the state-of-the-art methods on Argoverse 2 Motion Forecasting Leaderboard.}
	\resizebox{0.475\textwidth}{!}{
    \begin{tabular}{c|c|c|c|c|c}
    \hline
         \textbf{Methods} & avgB-MinFDE & avgMinADE &avgMinFDE  & actorMR & actorCR\\
         \hline
         LaneGCN & 3.90  & 1.49  & 3.24 & 0.37 & 0.07\\
         HiVT & 2.85  & 0.88  & 2.20 & 0.26 & 0.02 \\
         \rowcolor{gray!20} FFINet & \textbf{2.44} & \textbf{0.77} & \textbf{1.77} & \textbf{0.24}  & \textbf{0.02} \\
        \hline
    \end{tabular}}
    \label{tab:av2leaderboard}
\end{table}

\textbf{Comparison on Visualization.}
In Fig.~\ref{fig:Comparison}, we compare the visualization results of our FFINet with LaneGCN on the same scenarios of the Argoverse 1 validation set. Equipped with the data decomposition and feedback mechanism, our FFINet obtains more accurate predictions than LaneGCN in various traffic situations. 

\begin{figure}[t]
	\centering	\includegraphics[width=0.475\textwidth]{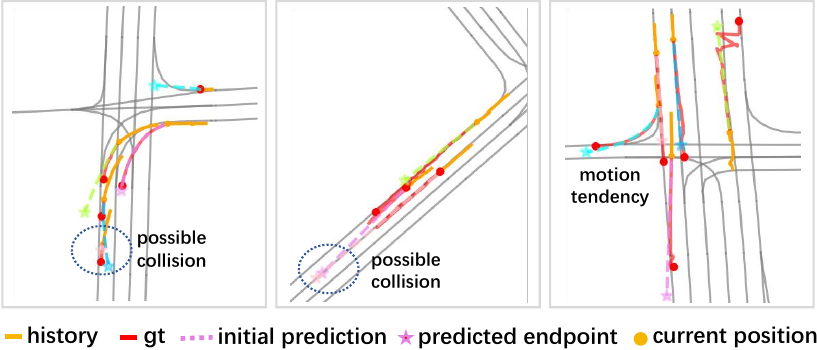}
    \caption{Visualization of initial trajectories. The first two columns reveal the possible collision between agents. The third column reflects the motion tendency of agents.}
	\label{fig:initial_collision}
\end{figure}

We also visualize the initial trajectories of multiple agents in Fig.~\ref{fig:initial_collision}. As we expected, the initial trajectories not only reflect the agent’s motion tendency, but also reveal the possible collision risks of agents. After encoding those initial trajectories, we aggregate the future features of neighbors into the features of predicted agent, so as to realize the encoding of future possible interaction.

\begin{figure*}[t]
	\centering
	\includegraphics[width=1.0\textwidth, height=0.2\textwidth]{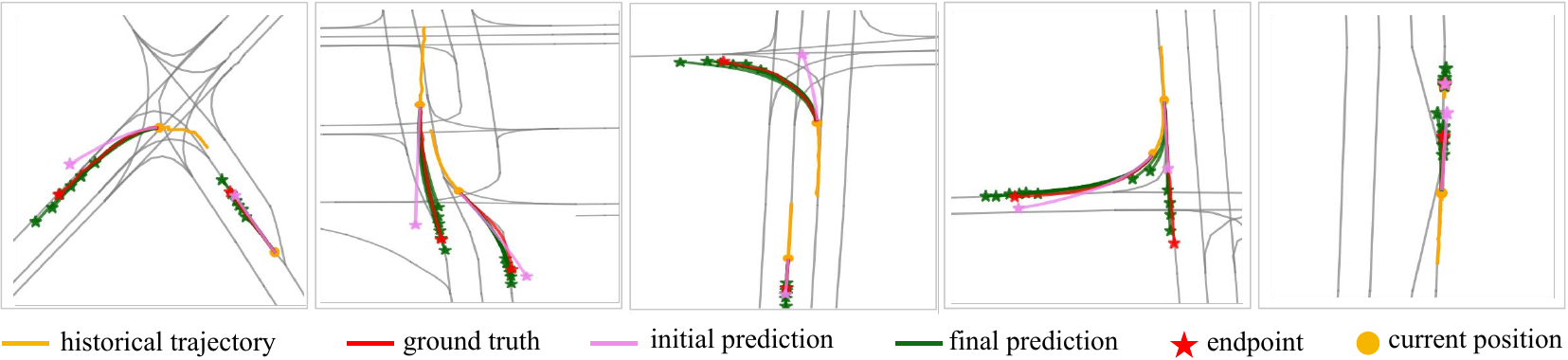}
	\caption{Visualization on joint prediction of multiple agents. For comparison of the initial prediction and final predictions, we represent them with a purple and green solid line, respectively.}
	\label{fig:initial}
\end{figure*}

Besides, we compare the results of FFINet with the corresponding initial prediction and final predictions in Fig.~\ref{fig:initial}. In particular, the first three columns show that the final predicted trajectories are more accurate than the initial predictions. The fourth column shows that the separated initial prediction of the successor and predecessor reasonably refine the final predictions at the crossroads. The predecessor regresses to the correct lane, and the successor accelerates straightly or turns right decreasingly. The last column illustrates a slowdown of the successor after previewing that the predecessor is almost motionless, which indicates that the feedback of future interactions can properly correct its feedforward prediction.

\subsection{Ablation Study}
We conduct various ablation studies on the validation set of Argoverse 1, in which metrics on $K=6$ is taken to measure the overall performance. Specifically, we first evaluate the effectiveness of each component of our FFINet, then do a comprehensive analysis of the data decomposition, the current fusion module and the future feedback interaction.

\textbf{Effectiveness of Each Component.} As shown in Table~\ref{tab:cp}, the results indicate that each component contributes to performance improvement of our FFINet. Firstly, the baseline model removes the current fusion module, future feedback interaction and global fusion module, which has seriously degraded the performance of our FFINet. Secondly, the current fusion module has the most significant impact on the performance, since the current topology information included in the current position coordinates plays an essential role in motion forecasting. Thirdly, without future feedback interaction, the model's performance drops dramatically, indicating that the feedback of future interactions is complemented by the observation interaction feature in motion prediction. Besides, the global fusion module can also noticeably improve the prediction performance.   

\begin{table}[]

    \footnotesize 
    \centering
    \caption{Ablation study results of different components, including Current Fusion Module~(Cur), Future Feedback Module~(Fut) and Global Fusion Module~(Glob).}
    \resizebox{0.475\textwidth}{!}{
    \begin{tabular}{c|c|c|c|c|c|c}
    \hline
         Cur & \hspace{0.1cm}Fut\hspace{0.1cm} & Glob  & b-minFDE & minADE & minFDE & MR\\
         \hline
         $\times$ & $\times$ &$\times$ & 1.618 & 0.667 & 0.938 & 0.082\\
        $\times$ & \checkmark & \checkmark & 1.594 & 0.650 & 0.915 & 0.077\\
        \checkmark & $\times$ &\checkmark 
        &1.604 & 0.656 & 0.926 & 0.080\\
        \checkmark & \checkmark & $\times$ & 1.602 & 0.654 & 0.925 & 0.079\\
        \rowcolor{gray!20}\checkmark & \checkmark & \checkmark & \textbf{1.584}& \textbf{0.650} & \textbf{0.905} & \textbf{0.076}\\
        \hline
    \end{tabular}}
    \label{tab:cp}
\end{table}
 
\textbf{Effectiveness of Data Decomposition.} To study the effects of representation decomposition, we remove the current absolute position fusion and the relative information aggregation in relative-correlation interaction blocks respectively. As shown in Table~\ref{tab:decompose}, without the current positions, the model cannot capture the current topology information and interactions of agents, resulting in degraded performance. Besides, the poor results of relative-correlation interaction blocks without relative information illustrate the effectiveness of relative information on interactions, which demonstrates the benefits of our data decomposition.

\begin{table}[]
    \footnotesize 
    \centering
    \caption{Ablation study results of data decomposition, including vectors, current position and relative information.}
	\resizebox{0.475\textwidth}{!}{
    \begin{tabular}{c|c|c|c|c|c|c}
    \hline
         Vector & Current position & Relatives & b-minFDE & minADE & minFDE & MR\\
         \hline
        \checkmark &$\times$ & \checkmark & 1.594 & 0.650 & 0.915 & 0.077 \\
         \checkmark & \checkmark & $\times$ & 2.012 & 0.800 & 1.331 & 0.164 \\
          \rowcolor{gray!20}\checkmark & \checkmark & \checkmark & 1.584 & 0.650 & 0.905 & 0.076\\
        \hline
    \end{tabular}} 
    \label{tab:decompose}
\end{table}

\textbf{Effectiveness of Current Fusion Module.} To have a deep look at our current fusion module, we conduct ablation experiments on the way of feature fusion, including fusion of neighbors in $G_{a2a}$, fusion of nearest and without any fusion. As shown in Table~\ref{tab:fuse}, the performance will be seriously degenerate by directly adding the current position features to corresponding vector features without any fusion operation, which indicates that the information of the near node plays a critical role in motion forecasting. Interestingly, we also note that fusing the information of the nearest node achieves better performance than fusing the neighbors. This suggests that the nearest information fusion in the current fusion module complements the interaction between neighbors and highlights the impact of the current timestamp on future prediction.

\begin{table}[]
    \footnotesize 
    \centering
    \caption{Ablation study results of Current Fusion Module, including fusion of neighbors(neighbor), fusion of nearest(nearest) and without fusion.}
     \resizebox{0.475\textwidth}{!}{
    \begin{tabular}{c|c|c|c|c|c}
    \hline
           neighbor & nearest & b-minFDE & minADE & minFDE & MR\\
         \hline
        $\times$ &$\times$ & 1.594 & 0.650 & 0.915 & 0.077\\
        \checkmark & $\times$ & 1.591 & 0.651 & 0.912& 0.077\\
        \rowcolor{gray!20}$\times$ & \checkmark & \textbf{1.584}& \textbf{0.650} & \textbf{0.905} & \textbf{0.076}\\
        \hline
    \end{tabular}}
    \label{tab:fuse}
\end{table}

\begin{table}[]
    \footnotesize 
    \centering
    \caption{Ablation study results of Future Feedback Module, including feedback(Back), future(Fut), and feedforward(Forward) interaction.}
    \resizebox{0.475\textwidth}{!}{
    \begin{tabular}{c|c|c|c|c|c|c}
    \hline
          Back&\hspace{0.2cm}Fut\hspace{0.5cm}&Forward& b-minFDE&minADE& minFDE & MR\\
         \hline
          $\times$ & $\times$ &$\times$ & 1.604 & 0.656 & 0.926 & 0.080\\
         \checkmark & $\times$ &$\times$ & 1.600 & 0.655 & 0.923 & 0.078\\
        \checkmark & \checkmark & $\times$ & 1.595 & 0.653 & 0.916& 0.078\\
       \rowcolor{gray!20}\checkmark & \checkmark & \checkmark & \textbf{1.584}& \textbf{0.650} & \textbf{0.905} & \textbf{0.076}\\
        \hline
    \end{tabular}}
    \label{tab:feedback}
\end{table}
\textbf{Effectiveness of Future Feedback Interaction.} To verify the effect of each interaction operator in the future feedback module, we conduct the ablation experiments on feedback interaction, future interaction and feedforward interaction, sequentially. As shown in Table~\ref{tab:feedback}, with the feedback interaction, all measure metrics are consistently improved, demonstrating the efficiency of feedback interaction. Besides, the performance is continually improved by integrating the possible interaction between agents in the future.  Finally, our model achieves the lowest distance error and miss rate, when it is equipped with the feedforward interaction.  

\begin{figure}[t]
	\centering
	\includegraphics[width=0.48\textwidth]{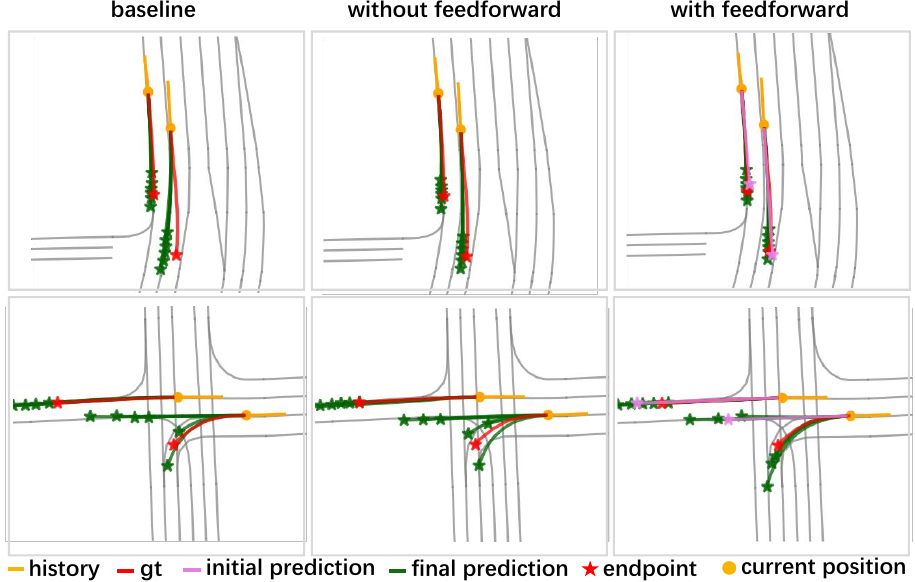}
	\caption{Visualizations of the baseline without three interaction modules, the baseline without future interaction feedback and the final model with future interaction feedback}
	\label{fig:feedforward}
\end{figure}

We also compare visualizations of the baseline without three interaction modules, the model without future interaction feedback and the final model with future interaction feedback in Fig.~\ref{fig:feedforward}. It indicates that our core future interaction feedback module can significantly improve prediction results.

\begin{table}[]
    \footnotesize 
    \centering
     \caption{Comparison on Computational cost with other methods.}
    \resizebox{0.475\textwidth}{!}{
    \begin{tabular}{c|c|c|c}
    \hline
         Method & Param(M) & FLOPs(G) & Infer time(ms)\\
         \hline
        LaneGCN & 3.681  & 2.011  & 2.662 \\
        DenseTNT(goal opt) & 0.618   & 0.552  & 85.852 \\
        DenseTNT(goal set pred) & 0.895 & 0.549 & 34.526 \\
        HiVT-128 & 2.287 & 0.317 & 2.058 \\
        \rowcolor{gray!20} FFINet & 6.200& 2.518 & 4.974 \\
        \hline
    \end{tabular}}
    \label{tab:cost}
\end{table}

\subsection{Analysis and Discussion}
For more comprehensive analysis of FFINet, we compare the efficiency of our model with other methods and evaluate the computation cost of each module of FFINet. Besides, we also discuss the influence of loss weights in this section.

\begin{table}[]
    \footnotesize 
    \centering
     \caption{Computational cost of each module, including Current Fusion Module (Cur), Future Feedback Module (Fut) and Global Fusion Module (Glob).}
    \resizebox{0.475\textwidth}{!}{
    \begin{tabular}{c|c|c|c|c|c}
    \hline
         Cur & Fut & Glob & Param(M) & FLOPs(G) & Inference time(ms)\\
         \hline
         $\times$ & $\times$ &$\times$ & 2.874  & 1.301 & 3.075 \\
        $\times$ & \checkmark & \checkmark & 5.701 &2.284  & 4.872  \\
        \checkmark & $\times$ &\checkmark 
        &4.873 & 2.247  & 4.336 \\
        \checkmark & \checkmark & $\times$ & 4.699  &  1.806&3.816 \\
        \rowcolor{gray!20}\checkmark & \checkmark & \checkmark & 6.200& 2.518 & 4.974 \\
        \hline
    \end{tabular}}
    \label{tab:eachfeedback}
\end{table}

\begin{table}[]
    \tiny 
    \centering
    \caption{Performance on different initial regression loss weight $\gamma$.}
    \resizebox{0.475\textwidth}{!}{
    \begin{tabular}{c|c|c|c|c}
    \hline
          $\gamma$& b-minFDE&minADE& minFDE & MR\\
         \hline
          0 & 1.599 & 0.657 & 0.921 & 0.0781 \\
          0.2 & \textbf{1.584} & \textbf{0.650} & \textbf{0.905} & \textbf{0.0756}\\
          0.4 & 1.587 & 0.652 & 0.908 & 0.0764 \\
          0.6 & 1.585 & \textbf{0.650} & 0.907 & 0.0766 \\
          0.8 & 1.599 & 0.655 & 0.922 & 0.0763 \\
          1.0 & 1.596 & 0.654 &0.919 & 0.0766 \\
        \hline
    \end{tabular}}
    \label{tab:gamma}
\end{table}

\begin{table}[]
    \tiny 
    \centering
    \caption{Performance on different endpoint loss weights $\lambda$.}
    \resizebox{0.475\textwidth}{!}{
    \begin{tabular}{c|c|c|c|c}
    \hline
          $\lambda$& b-minFDE&minADE& minFDE & MR\\
         \hline
          0 & 1.642 & 0.666 & 0.966 & 0.0856  \\
          0.1 & 1.604 & 0.653  & 0.927  & 0.0798 \\
          0.3 & 1.595 &  0.653 & 0.918 & 0.0769 \\
          0.5 & \textbf{1.584} & \textbf{0.650} & \textbf{0.905} & 0.0756\\
          0.7& 1.593  & 0.653 & 0.914 & \textbf{0.0752}  \\
          0.9& 1.602 & 0.657  & 0.924  & 0.0788 \\
        \hline
    \end{tabular}}
    \label{tab:lambda}
\end{table}

\textbf{Computational Cost.}~We compare the parameters, FLOPs (Floating Point Operations) and average inference runtime per scenario with other methods on the same devices. As shown in Table \ref{tab:cost}, we have very close inference time with LaneGCN and HiVT. Our FFINet is faster than DenseTNT by a large margin and achieves better performance than the comparison methods, which means our model achieves considerable performance
improvement with limited computation cost.

We also evaluate the parameters, FLOPs and average inference runtime per scenario of each module of FFINet. As shown in Table \ref{tab:eachfeedback}, each module gains considerable performance with very limited computational cost. Equipped with three interaction blocks, our model still keeps the comparative inference time compared with the base model.

\textbf{Performance on Different Loss Weights.}
For exploring the influence of loss weight, we trained various models with different weights and evaluated them on the Argoverse 1 validation set. As shown in Table~\ref{tab:gamma}, we illustrate the performance of models with various $\gamma$ and fixed $\lambda=0.5, \beta=2.0$ on the metrics of K=6. Without the initial regression loss, the performance degrades dramatically on all metrics simultaneously. Equipped with the initial regression loss, the performance of models is enhanced to a certain extent. Besides, the model achieves the best performance with $\gamma=0.2$. 

Similarity, Table~\ref{tab:lambda} shows the performance of models with diverse $\lambda$, where $\gamma=0.2, \beta=2.0$. With $\lambda=0.5$, the model achieves the best performance on distance metrics, while the best performance of MR is obtained with $\lambda=0.7$. The endpoint loss can improve the performance of distance error and miss rate significantly.

\section{Conclusion}
\label{Conclusion}
In this paper, we propose a novel Future Feedback Interaction Network~(FFINet) for motion forecasting, where current, historical and future interaction features are integrated effectively.
With the current fusion module, our model emphasizes the scene topology information at the current timestamp, which in turn improves the prediction performance. Besides, the future feedback module is proposed to extract the future interactions and feed the
plausible future interactions of agents back to the observation interaction features, which dramatically reduces the collision rate on multi-agent predictions.
Extensive experiments on Argoverse 1 and Argoverse 2 benchmarks indicate that our method outperforms most of the state-of-the-art methods both on single-agent and multi-agent prediction. Furthermore, the proposed plug-in relative interaction block can be easily integrated with other interaction models.

Although our FFINet achieves promising results with the current positions, our model loses the scene translation-invariance and only keeps the rotation-invariance. To our best knowledge, data augmentation, such as scene translation and trajectory scaling, can alleviate this problem to some degree. We leave it as future work to explore the influence of scene translation in our model.

\ifCLASSOPTIONcaptionsoff
  \newpage
\fi

{
\bibliographystyle{IEEEtran}
\bibliography{IEEEabrv,egbib}

\begin{thebibliography}{10}
\providecommand{\url}[1]{#1}
\csname url@samestyle\endcsname
\providecommand{\newblock}{\relax}
\providecommand{\bibinfo}[2]{#2}
\providecommand{\BIBentrySTDinterwordspacing}{\spaceskip=0pt\relax}
\providecommand{\BIBentryALTinterwordstretchfactor}{4}
\providecommand{\BIBentryALTinterwordspacing}{\spaceskip=\fontdimen2\font plus
\BIBentryALTinterwordstretchfactor\fontdimen3\font minus \fontdimen4\font\relax}
\providecommand{\BIBforeignlanguage}[2]{{%
\expandafter\ifx\csname l@#1\endcsname\relax
\typeout{** WARNING: IEEEtran.bst: No hyphenation pattern has been}%
\typeout{** loaded for the language `#1'. Using the pattern for}%
\typeout{** the default language instead.}%
\else
\language=\csname l@#1\endcsname
\fi
#2}}
\providecommand{\BIBdecl}{\relax}
\BIBdecl

\bibitem{helbing1995social}
D.~Helbing and P.~Molnar, ``Social force model for pedestrian dynamics,'' \emph{Physical review E}, vol.~51, no.~5, p. 4282, 1995.

\bibitem{varadarajan2022multipath++}
B.~Varadarajan, A.~Hefny, A.~Srivastava, K.~S. Refaat, N.~Nayakanti, A.~Cornman, K.~Chen, B.~Douillard, C.~P. Lam, D.~Anguelov \emph{et~al.}, ``Multipath++: Efficient information fusion and trajectory aggregation for behavior prediction,'' in \emph{2022 International Conference on Robotics and Automation (ICRA)}.\hskip 1em plus 0.5em minus 0.4em\relax IEEE, 2022, pp. 7814--7821.

\bibitem{zeng2021lanercnn}
W.~Zeng, M.~Liang, R.~Liao, and R.~Urtasun, ``Lanercnn: Distributed representations for graph-centric motion forecasting,'' in \emph{2021 IEEE/RSJ International Conference on Intelligent Robots and Systems (IROS)}.\hskip 1em plus 0.5em minus 0.4em\relax IEEE, 2021, pp. 532--539.

\bibitem{shi2021sgcn}
L.~Shi, L.~Wang, C.~Long, S.~Zhou, M.~Zhou, Z.~Niu, and G.~Hua, ``Sgcn: Sparse graph convolution network for pedestrian trajectory prediction,'' in \emph{Proceedings of the IEEE/CVF Conference on Computer Vision and Pattern Recognition}, 2021, pp. 8994--9003.

\bibitem{gao2020vectornet}
J.~Gao, C.~Sun, H.~Zhao, Y.~Shen, D.~Anguelov, C.~Li, and C.~Schmid, ``Vectornet: Encoding hd maps and agent dynamics from vectorized representation,'' in \emph{Proceedings of the IEEE/CVF Conference on Computer Vision and Pattern Recognition}, 2020, pp. 11\,525--11\,533.

\bibitem{gu2021densetnt}
J.~Gu, C.~Sun, and H.~Zhao, ``Densetnt: End-to-end trajectory prediction from dense goal sets,'' in \emph{Proceedings of the IEEE/CVF International Conference on Computer Vision}, 2021, pp. 15\,303--15\,312.

\bibitem{li2020end}
L.~L. Li, B.~Yang, M.~Liang, W.~Zeng, M.~Ren, S.~Segal, and R.~Urtasun, ``End-to-end contextual perception and prediction with interaction transformer,'' in \emph{2020 IEEE/RSJ International Conference on Intelligent Robots and Systems (IROS)}.\hskip 1em plus 0.5em minus 0.4em\relax IEEE, 2020, pp. 5784--5791.

\bibitem{tang2019multiple}
C.~Tang and R.~R. Salakhutdinov, ``Multiple futures prediction,'' \emph{Advances in Neural Information Processing Systems}, vol.~32, 2019.

\bibitem{casas2020implicit}
S.~Casas, C.~Gulino, S.~Suo, K.~Luo, R.~Liao, and R.~Urtasun, ``Implicit latent variable model for scene-consistent motion forecasting,'' in \emph{European Conference on Computer Vision}.\hskip 1em plus 0.5em minus 0.4em\relax Springer, 2020, pp. 624--641.

\bibitem{sun2022m2i}
Q.~Sun, X.~Huang, J.~Gu, B.~C. Williams, and H.~Zhao, ``M2i: From factored marginal trajectory prediction to interactive prediction,'' in \emph{Proceedings of the IEEE/CVF Conference on Computer Vision and Pattern Recognition}, 2022, pp. 6543--6552.

\bibitem{song2022learning}
H.~Song, D.~Luan, W.~Ding, M.~Y. Wang, and Q.~Chen, ``Learning to predict vehicle trajectories with model-based planning,'' in \emph{Conference on Robot Learning}.\hskip 1em plus 0.5em minus 0.4em\relax PMLR, 2022, pp. 1035--1045.

\bibitem{gilles2022gohome}
T.~Gilles, S.~Sabatini, D.~Tsishkou, B.~Stanciulescu, and F.~Moutarde, ``Gohome: Graph-oriented heatmap output for future motion estimation,'' in \emph{2022 International Conference on Robotics and Automation (ICRA)}.\hskip 1em plus 0.5em minus 0.4em\relax IEEE, 2022, pp. 9107--9114.

\bibitem{nash2016review}
C.~J. Nash, D.~J. Cole, and R.~S. Bigler, ``A review of human sensory dynamics for application to models of driver steering and speed control,'' \emph{Biological cybernetics}, vol. 110, no.~2, pp. 91--116, 2016.

\bibitem{antonini2006discrete}
G.~Antonini, M.~Bierlaire, and M.~Weber, ``Discrete choice models of pedestrian walking behavior,'' \emph{Transportation Research Part B: Methodological}, vol.~40, no.~8, pp. 667--687, 2006.

\bibitem{messaoud2019relational}
K.~Messaoud, I.~Yahiaoui, A.~Verroust-Blondet, and F.~Nashashibi, ``Relational recurrent neural networks for vehicle trajectory prediction,'' in \emph{2019 IEEE Intelligent Transportation Systems Conference (ITSC)}.\hskip 1em plus 0.5em minus 0.4em\relax IEEE, 2019, pp. 1813--1818.

\bibitem{lee2022muse}
M.~Lee, S.~S. Sohn, S.~Moon, S.~Yoon, M.~Kapadia, and V.~Pavlovic, ``Muse-vae: Multi-scale vae for environment-aware long term trajectory prediction,'' in \emph{Proceedings of the IEEE/CVF Conference on Computer Vision and Pattern Recognition}, 2022, pp. 2221--2230.

\bibitem{alahi2016social}
A.~Alahi, K.~Goel, V.~Ramanathan, A.~Robicquet, L.~Fei-Fei, and S.~Savarese, ``Social lstm: Human trajectory prediction in crowded spaces,'' in \emph{Proceedings of the IEEE conference on computer vision and pattern recognition}, 2016, pp. 961--971.

\bibitem{li2020evolvegraph}
J.~Li, F.~Yang, M.~Tomizuka, and C.~Choi, ``Evolvegraph: Multi-agent trajectory prediction with dynamic relational reasoning,'' \emph{Advances in neural information processing systems}, vol.~33, pp. 19\,783--19\,794, 2020.

\bibitem{liang2020learning}
M.~Liang, B.~Yang, R.~Hu, Y.~Chen, R.~Liao, S.~Feng, and R.~Urtasun, ``Learning lane graph representations for motion forecasting,'' in \emph{European Conference on Computer Vision}.\hskip 1em plus 0.5em minus 0.4em\relax Springer, 2020, pp. 541--556.

\bibitem{zhou2022hivt}
Z.~Zhou, L.~Ye, J.~Wang, K.~Wu, and K.~Lu, ``Hivt: Hierarchical vector transformer for multi-agent motion prediction,'' in \emph{Proceedings of the IEEE/CVF Conference on Computer Vision and Pattern Recognition}, 2022, pp. 8823--8833.

\bibitem{girgis2021latent}
R.~Girgis, F.~Golemo, F.~Codevilla, M.~Weiss, J.~A. D'Souza, S.~E. Kahou, F.~Heide, and C.~Pal, ``Latent variable sequential set transformers for joint multi-agent motion prediction,'' \emph{arXiv preprint arXiv:2104.00563}, 2021.

\bibitem{ngiam2021scene}
J.~Ngiam, B.~Caine, V.~Vasudevan, Z.~Zhang, H.-T.~L. Chiang, J.~Ling, R.~Roelofs, A.~Bewley, C.~Liu, A.~Venugopal \emph{et~al.}, ``Scene transformer: A unified multi-task model for behavior prediction and planning,'' \emph{arXiv e-prints}, pp. arXiv--2106, 2021.

\bibitem{jia2022multi}
X.~Jia, L.~Sun, H.~Zhao, M.~Tomizuka, and W.~Zhan, ``Multi-agent trajectory prediction by combining egocentric and allocentric views,'' in \emph{Conference on Robot Learning}.\hskip 1em plus 0.5em minus 0.4em\relax PMLR, 2022, pp. 1434--1443.

\bibitem{kofinas2021roto}
M.~Kofinas, N.~Nagaraja, and E.~Gavves, ``Roto-translated local coordinate frames for interacting dynamical systems,'' \emph{Advances in Neural Information Processing Systems}, vol.~34, pp. 6417--6429, 2021.

\bibitem{shi2022motion}
S.~Shi, L.~Jiang, D.~Dai, and B.~Schiele, ``Motion transformer with global intention localization and local movement refinement,'' \emph{Advances in Neural Information Processing Systems}, vol.~35, pp. 6531--6543, 2022.

\bibitem{tolstaya2021identifying}
E.~Tolstaya, R.~Mahjourian, C.~Downey, B.~Vadarajan, B.~Sapp, and D.~Anguelov, ``Identifying driver interactions via conditional behavior prediction,'' in \emph{2021 IEEE International Conference on Robotics and Automation (ICRA)}.\hskip 1em plus 0.5em minus 0.4em\relax IEEE, 2021, pp. 3473--3479.

\bibitem{mangalam2020not}
K.~Mangalam, H.~Girase, S.~Agarwal, K.-H. Lee, E.~Adeli, J.~Malik, and A.~Gaidon, ``It is not the journey but the destination: Endpoint conditioned trajectory prediction,'' in \emph{European conference on computer vision}.\hskip 1em plus 0.5em minus 0.4em\relax Springer, 2020, pp. 759--776.

\bibitem{wang2023ganet}
M.~Wang, X.~Zhu, C.~Yu, W.~Li, Y.~Ma, R.~Jin, X.~Ren, D.~Ren, M.~Wang, and W.~Yang, ``Ganet: Goal area network for motion forecasting,'' in \emph{2023 IEEE International Conference on Robotics and Automation (ICRA)}.\hskip 1em plus 0.5em minus 0.4em\relax IEEE, 2023, pp. 1609--1615.

\bibitem{xu2022remember}
C.~Xu, W.~Mao, W.~Zhang, and S.~Chen, ``Remember intentions: Retrospective-memory-based trajectory prediction,'' in \emph{Proceedings of the IEEE/CVF Conference on Computer Vision and Pattern Recognition}, 2022, pp. 6488--6497.

\bibitem{chen2022scept}
Y.~Chen, B.~Ivanovic, and M.~Pavone, ``Scept: Scene-consistent, policy-based trajectory predictions for planning,'' in \emph{Proceedings of the IEEE/CVF Conference on Computer Vision and Pattern Recognition}, 2022, pp. 17\,103--17\,112.

\bibitem{mangalam2021goals}
K.~Mangalam, Y.~An, H.~Girase, and J.~Malik, ``From goals, waypoints \& paths to long term human trajectory forecasting,'' in \emph{Proceedings of the IEEE/CVF International Conference on Computer Vision}, 2021, pp. 15\,233--15\,242.

\bibitem{zhao2020tnt}
H.~Zhao, J.~Gao, T.~Lan, C.~Sun, B.~Sapp, B.~Varadarajan, Y.~Shen, Y.~Shen, Y.~Chai, C.~Schmid \emph{et~al.}, ``Tnt: Target-driven trajectory prediction,'' \emph{arXiv preprint arXiv:2008.08294}, 2020.

\bibitem{lee2016stochastic}
S.~Lee, S.~Purushwalkam Shiva~Prakash, M.~Cogswell, V.~Ranjan, D.~Crandall, and D.~Batra, ``Stochastic multiple choice learning for training diverse deep ensembles,'' \emph{Advances in Neural Information Processing Systems}, vol.~29, 2016.

\bibitem{ye2021tpcn}
M.~Ye, T.~Cao, and Q.~Chen, ``Tpcn: Temporal point cloud networks for motion forecasting,'' in \emph{Proceedings of the IEEE/CVF Conference on Computer Vision and Pattern Recognition}, 2021, pp. 11\,318--11\,327.

\bibitem{zhang2022trajectory}
L.~Zhang, P.~Li, J.~Chen, and S.~Shen, ``Trajectory prediction with graph-based dual-scale context fusion,'' 2022.

\bibitem{wang2022tenet}
Y.~Wang, H.~Zhou, Z.~Zhang, C.~Feng, H.~Lin, C.~Gao, Y.~Tang, Z.~Zhao, S.~Zhang, J.~Guo \emph{et~al.}, ``Tenet: Transformer encoding network for effective temporal flow on motion prediction,'' \emph{arXiv preprint arXiv:2207.00170}, 2022.

\bibitem{nayakanti2023wayformer}
N.~Nayakanti, R.~Al-Rfou, A.~Zhou, K.~Goel, K.~S. Refaat, and B.~Sapp, ``Wayformer: Motion forecasting via simple \& efficient attention networks,'' in \emph{2023 IEEE International Conference on Robotics and Automation (ICRA)}.\hskip 1em plus 0.5em minus 0.4em\relax IEEE, 2023, pp. 2980--2987.

\bibitem{gilles2022thomas}
T.~Gilles, S.~Sabatini, D.~Tsishkou, B.~Stanciulescu, and F.~Moutarde, ``Thomas: Trajectory heatmap output with learned multi-agent sampling,'' in \emph{International Conference on Learning Representations}, 2022.

\bibitem{cui2023gorela}
A.~Cui, S.~Casas, K.~Wong, S.~Suo, and R.~Urtasun, ``Gorela: Go relative for viewpoint-invariant motion forecasting,'' in \emph{2023 IEEE International Conference on Robotics and Automation (ICRA)}.\hskip 1em plus 0.5em minus 0.4em\relax IEEE, 2023, pp. 7801--7807.

\bibitem{park2023leveraging}
D.~Park, H.~Ryu, Y.~Yang, J.~Cho, J.~Kim, and K.-J. Yoon, ``Leveraging future relationship reasoning for vehicle trajectory prediction,'' in \emph{International Conference on Learning Representations (ICLR 2023)}.\hskip 1em plus 0.5em minus 0.4em\relax Eleventh International Conference on Learning Representations, 2023.

\bibitem{argoverse2022comp}
``Argoverse motion forecasting competition,'' \url{https://eval.ai/web/challenges/challenge-page/454/overview}, Accessed: 2022-11-11.

\bibitem{argoverse22023comp}
``Argoverse2: Motion forecasting competition,'' \url{https://eval.ai/web/challenges/challenge-page/1719/overview}, Accessed: 2023-08-03.

\bibitem{cui2019multimodal}
H.~Cui, V.~Radosavljevic, F.-C. Chou, T.-H. Lin, T.~Nguyen, T.-K. Huang, J.~Schneider, and N.~Djuric, ``Multimodal trajectory predictions for autonomous driving using deep convolutional networks,'' in \emph{2019 International Conference on Robotics and Automation (ICRA)}.\hskip 1em plus 0.5em minus 0.4em\relax IEEE, 2019, pp. 2090--2096.

\bibitem{chai2020multipath}
Y.~Chai, B.~Sapp, M.~Bansal, and D.~Anguelov, ``Multipath: Multiple probabilistic anchor trajectory hypotheses for behavior prediction,'' in \emph{Conference on Robot Learning}.\hskip 1em plus 0.5em minus 0.4em\relax PMLR, 2020, pp. 86--99.

\end{thebibliography}
}

\begin{IEEEbiography}[{\includegraphics[width=0.95in,height=1.25in,clip]{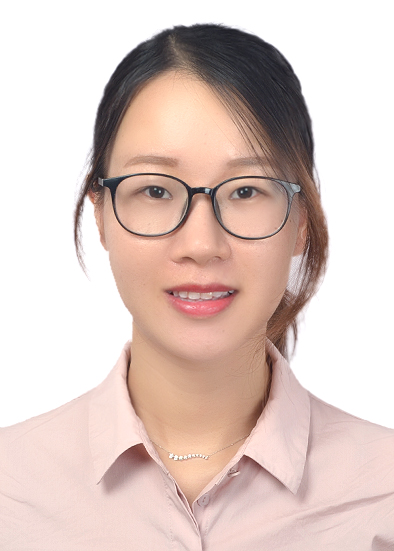}}]{Miao Kang}
received the M.S. degree in photogrammetry and remote sensing in 2017  from  National University of Defense Technology and is currently working toward the Ph.D. degree with the Institute of Artificial Intelligence and Robotics at  Xi'an Jiaotong University, Xi'an, China. Her research interests include deep learning, computer vision and autonomous driving, with a focus on object detection and trajectory prediction.
\end{IEEEbiography}

\begin{IEEEbiography}[{\includegraphics[width=0.95in,height=1.25in,clip]{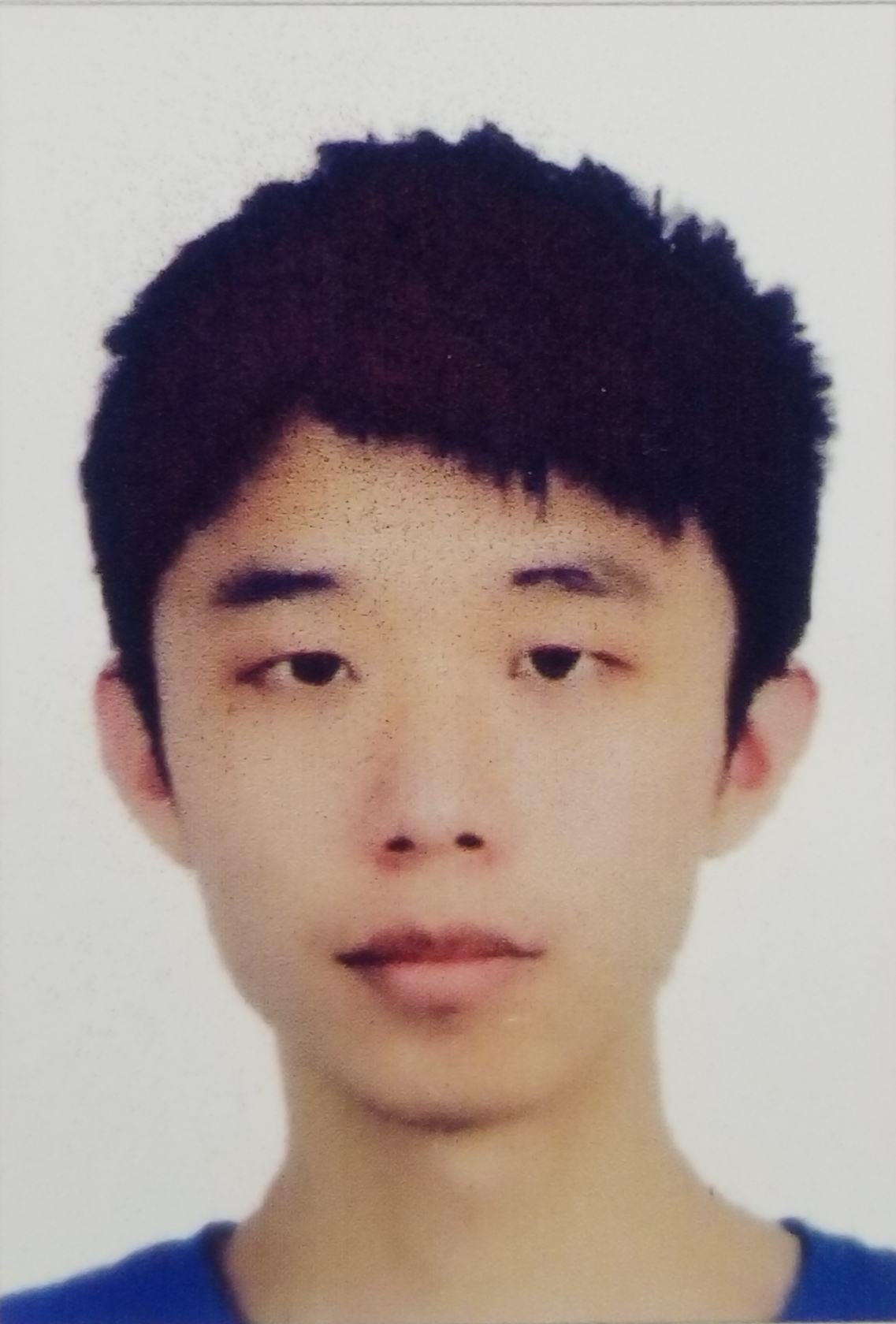}}]{Shengqi Wang} received his Master degree from the Institute of Artificial Intelligence and Robotics, Xi'an Jiaotong University in 2022. His research interest includes computer vision, motion forecasting and geometric deep learning.
\end{IEEEbiography}

\begin{IEEEbiography}[{\includegraphics[width=0.95in,height=1.25in,clip]{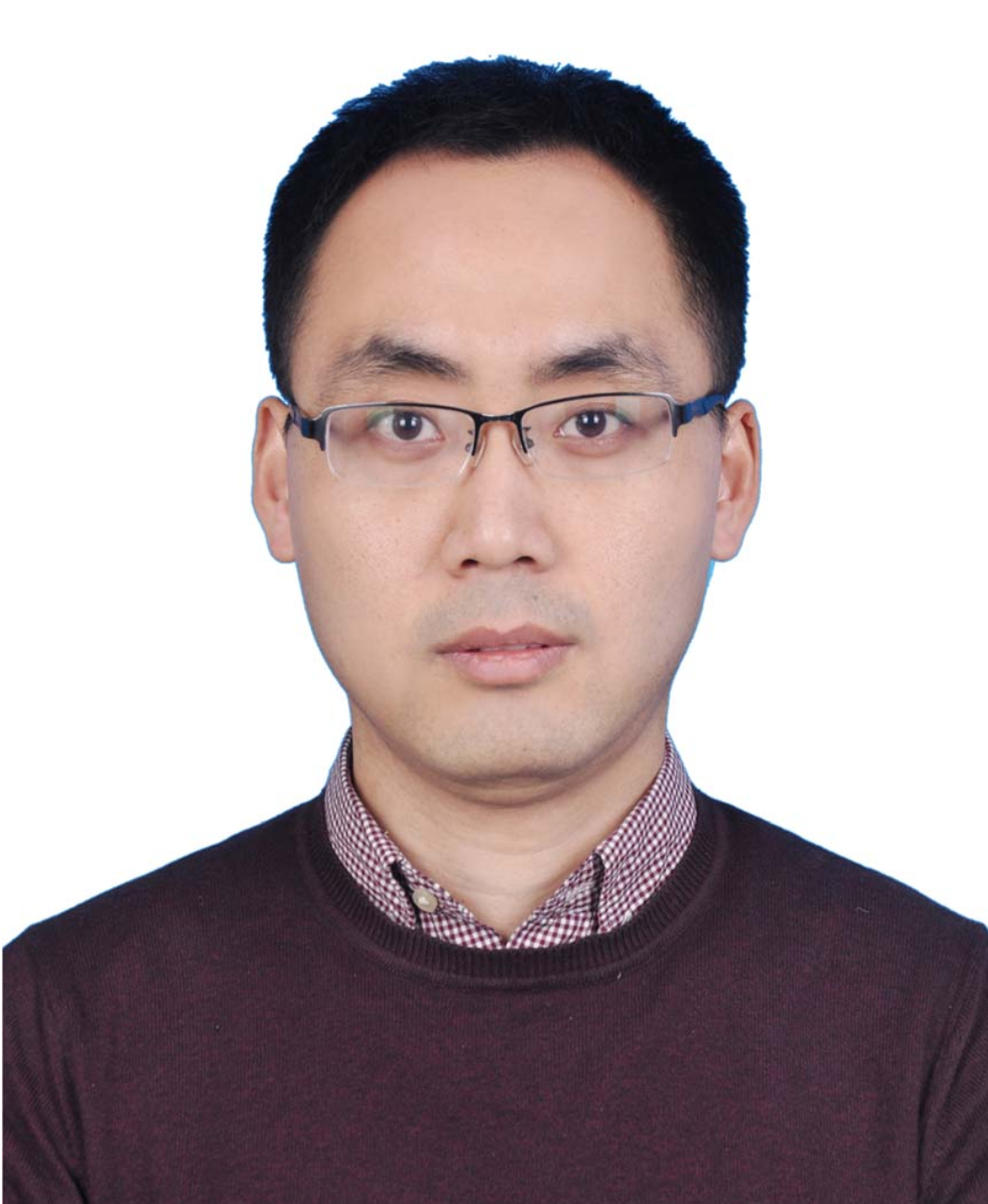}}]{Sanping Zhou}
(Member, IEEE) received his PhD. degree from Xi'an Jiaotong University, Xi'an, China, in 2020. From 2018 to 2019, he was a Visiting Ph.D. Student with Robotics Institute, Carnegie Mellon University.  He is currently an Assistant Professor with the Institute of Artificial Intelligence and Robotics at Xi'an Jiaotong University. His research interests include machine learning, deep learning and computer vision, with a focus on medical image segmentation, person re-identification, salient object detection, image classification and visual tracking.
\end{IEEEbiography}

\begin{IEEEbiography}[{\includegraphics[width=0.95in,height=1.25in,clip]{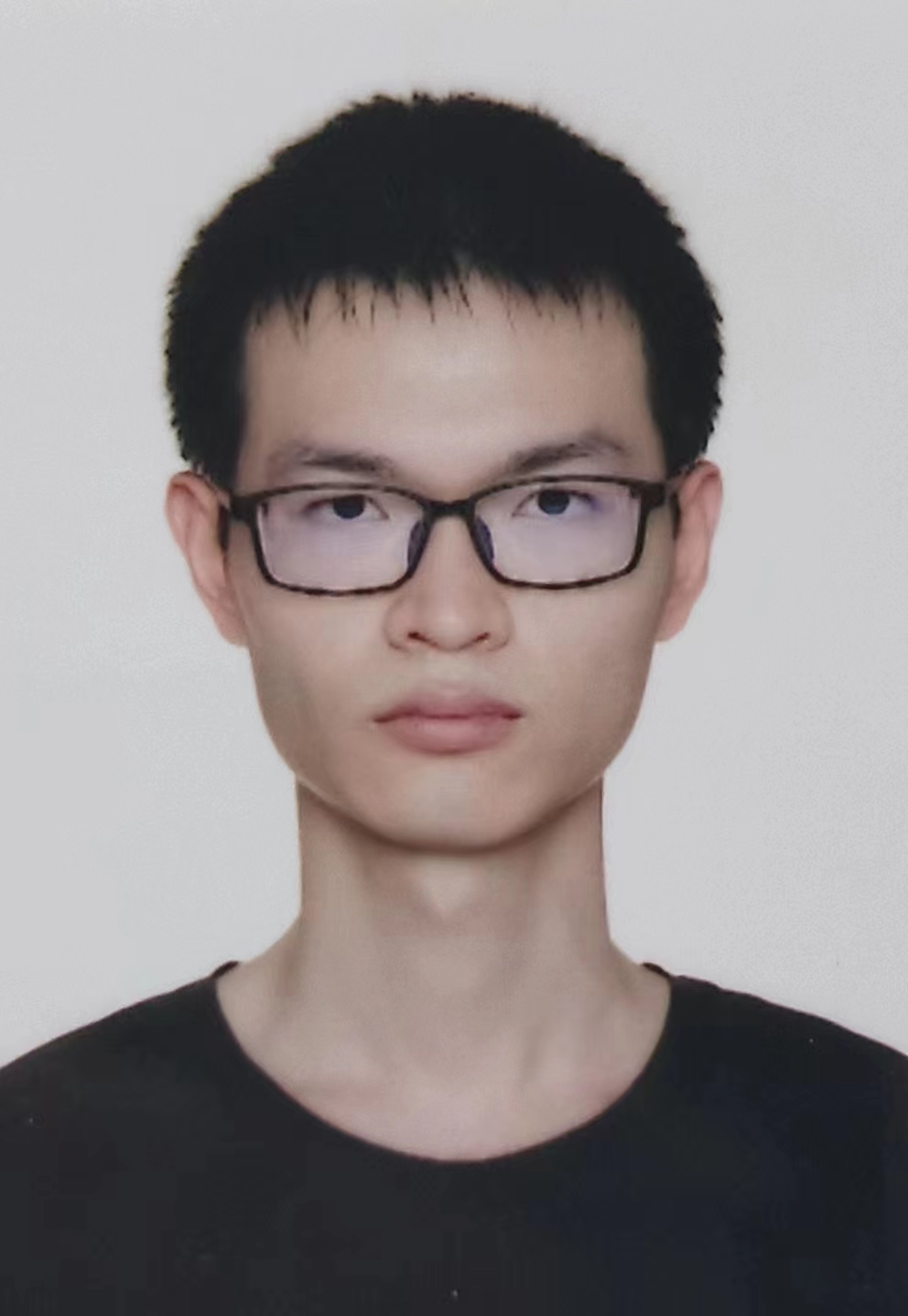}}]{Ke Ye} is working towards the Master degree in In Institute of Artificial Intelligence and Robotics at Xi’an Jiaotong University. He received his B.E. degree in Artificial Intelligence from Xi’an Jiaotong University in 2022. His current research interest focuses on trajectory prediction.
\end{IEEEbiography}

\begin{IEEEbiography}[{\includegraphics[width=0.95in,height=1.25in,clip]{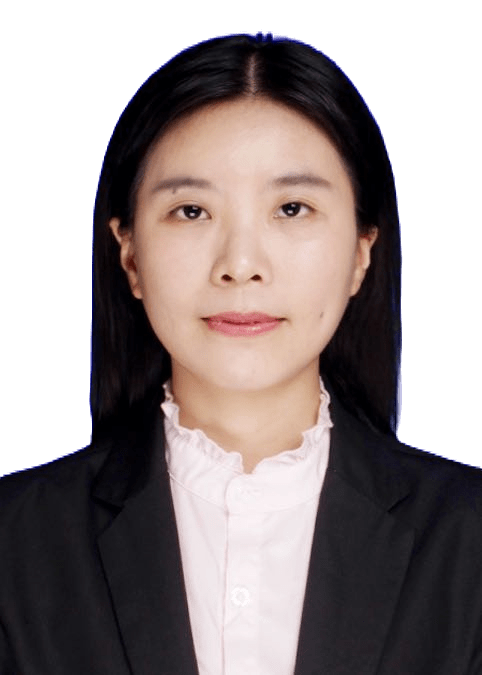}}]{Jingjing Jiang} {Jingjing Jiang} received her Ph.D. degree from the Institute of Artificial Intelligence and Robotics, Xi'an Jiaotong University in 2023. Her research interest lies at the intersection of computer vision and natural language processing, including visual question answering, image captioning, and multimodal model adaption.
\end{IEEEbiography}
\vspace{-15.2cm}
\begin{IEEEbiography}[{\includegraphics[width=0.95in,height=1.25in,clip,keepaspectratio]{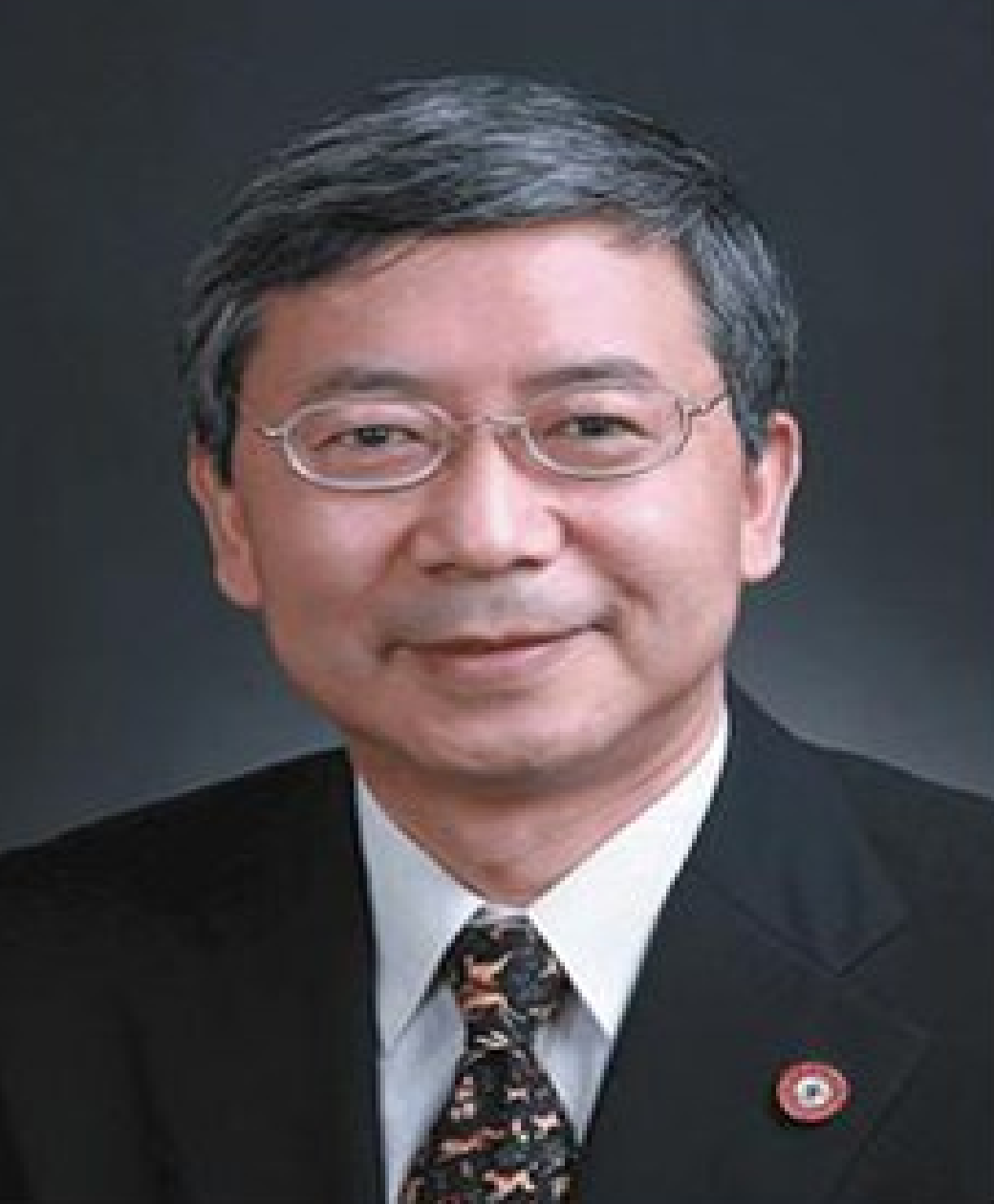}}]{Nanning Zheng}
(IEEE SM'93-F'06) received the Graduate degree in electrical engineering and the M.S. degree in information and control engineering from Xi'an Jiaotong University, Xi'an, China, in 1975 and 1981, respectively, and the Ph.D. degree in electrical engineering from Keio University, Yokohama, Japan, in 1985.
He is a Distinguished Professor with the Institute of Artificial Intelligence and Robotics, Xi'an Jiaotong University.
He is the Founder of the Institute of Artificial Intelligence and Robotics, Xi'an Jiaotong University.
He is currently a Professor and Director of the Institute of Artificial Intelligence and Robotics, Xi'an Jiaotong University. His research interests include computer vision, pattern recognition, autonomous vehicle, and brain-inspired computing.
He became a member of the Chinese Academy of Engineering in 1999.
He is a Council Member of the International Association for Pattern Recognition.
\end{IEEEbiography}

\end{document}